\documentclass[lettersize,journal]{IEEEtran}
\usepackage{amsmath,amsfonts}
\usepackage{algorithmic}
\usepackage{algorithm}
\usepackage{array}
\usepackage{subcaption}
\usepackage{tcolorbox}
\usepackage{textcomp}
\usepackage{stfloats}
\usepackage{url}
\usepackage{verbatim}
\usepackage{graphicx}
\usepackage{caption}
\usepackage{cite}
\usepackage{amssymb}
\usepackage{listings}
\usepackage{xcolor}
\usepackage{multicol}

\usepackage[hidelinks,bookmarks=true]{hyperref}
\usepackage{amsmath}
\usepackage[capitalise]{cleveref}
\usepackage[para]{footmisc}
\usepackage{xcolor}
\usepackage{graphicx}
\usepackage{booktabs}
\usepackage{wrapfig}
\usepackage{lipsum}
\usepackage{float}
\usepackage{enumitem}
\usepackage{lipsum}
\usepackage{adjustbox}
\usepackage{graphicx}
\usepackage{subcaption}
\usepackage{siunitx}
\usepackage{pifont}
\usepackage[utf8]{inputenc} 
\usepackage{amssymb}
\usepackage{bm}
\usepackage{placeins}
\usepackage{tikz}
\usepackage{glossaries}
\usepackage{multirow}
\usepackage{makecell}
\usepackage{lipsum}
\usepackage{cuted}
\usepackage{tabularx}

\newcommand{\cmark}{\ding{51}}%
\newcommand{\xmark}{\ding{55}} 

\usepackage{amssymb}

\usepackage{listings}
\usepackage{xcolor}
\definecolor{localnav}{HTML}{534ace}
\definecolor{purplei}{HTML}{9673A6}
\definecolor{yellowi}{HTML}{D6B656}
\definecolor{triangle}{HTML}{FA6800}
\definecolor{greeni}{HTML}{3A5431}
\definecolor{redi}{HTML}{B20000}

\title{\textcolor{localnav}{LocalNav:} Distilling Frontier VLMs and Embodied RL for On-Device Object Goal Navigation}

\author{
    Nicolas Baumann$^{1, 2, *}$,
    Liam Boyle$^{1, *}$,
    Pu Deng$^{2}$,
    Edoardo Ghignone$^{1}$,
    Boyang Sun$^{3}$,\\
    Marc Pollefeys$^{3}$,
    Luca Benini$^{2}$,
    Michele Magno$^{1}$\\
    \small $^{1}$Center for Project-Based Learning, $^{2}$Integrated Systems Laboratory, $^{3}$Computer Vision and Geometry\\
    \small ETH Zurich
}

\newacronym{llm}{LLM}{Large Language Model}
\newacronym{vlm}{VLM}{Vision Language Model}
\newacronym{vla}{VLA}{Vision Language Action}
\newacronym{rag}{RAG}{Retrieval Augmented Generation}
\newacronym{lora}{LoRA}{Low Rank Adaptation}
\newacronym{mpc}{MPC}{Model Predictive Controller}
\newacronym{rmse}{RMSE}{Root Mean Square Error}
\newacronym{hmi}{HMI}{Human Machine Interaction}
\newacronym{ads}{ADS}{Autonomous Driving Systems}
\newacronym{ml}{ML}{Machine Learning}
\newacronym{nn}{NN}{Neural Network}
\newacronym{rl}{RL}{Reinforcement Learning}
\newacronym{ai}{AI}{Artificial Intelligence}
\newacronym{obc}{OBC}{OnBoard Computer}
\newacronym{gpu}{GPU}{Graphics Processing Unit}
\newacronym{cpu}{CPU}{Central Processing Unit}
\newacronym{ram}{RAM}{Random Access Memory}
\newacronym{ros}{ROS}{Robot Operating System}
\newacronym{peft}{PEFT}{Parameter Efficient Fine-Tuning}
\newacronym{cot}{CoT}{Chain of Thought}
\newacronym{sft}{SFT}{Supervised Fine-Tuning}
\newacronym{grpo}{GRPO}{Group Relative Policy Optimization}
\newacronym{eai}{Embodied AI}{Embodied Artificial Intelligence}
\newacronym{sota}{SotA}{State of the Art}
\newacronym{rlvr}{RLVR}{Reinforcement Learning from Verifiable Rewards}
\newacronym{erlvr}{E-RLVR}{Embodied Reinforcement Learning from Verifiable Rewards}
\newacronym{sr}{SR}{Success Rate}
\newacronym{spl}{SPL}{Success weighted by Path Length}
\newacronym{hm3d}{HM3D}{Habitat-Matterport 3D Research Dataset}
\newacronym{mp3d}{MP3D}{MatterPort 3D}
\newacronym{mlp}{MLP}{Multi Layer Perceptron}
\newacronym{ovon}{OVON}{Open-Vocabulary Object goal Navigation}
\newacronym{sg}{SG}{Scene Graph}
\newacronym{slam}{SLAM}{Simultaneous Localization and Mapping}
\newacronym{ogn}{ObjectNav}{Object Goal Navigation}
\newacronym{bc}{BC}{Behavior Cloning}
\newacronym{vio}{VIO}{Visual-Inertial Odometry}
\newacronym{fov}{FoV}{Field of View}
\newacronym{ttc}{TTC}{Test-Time Computation}
\newacronym{ddppo}{DD-PPO}{Decentralized Distributed Proximal Policy Optimization}
\newacronym{sru}{SRU}{Spatially-Enhanced Recurrent Units}
\newacronym{vae}{VAE}{Variational AutoEncoder}
\newacronym{rnn}{RNN}{Recurrent Neural Network}
\newacronym{dcdc}{DC/DC}{Direct Current to Direct Current}
\newacronym{cad}{CAD}{Computer-Aided Design}
\newacronym{bom}{BoM}{Bill of Materials}
\newacronym{KLD}{KLD}{Kullback–Leibler divergence}
\newacronym{pp}{PP}{Prompt Processing}
\newacronym{tg}{TG}{Token Generation}

\begin{document}
\bstctlcite{IEEEexample:BSTcontrol}
\maketitle

\begin{strip}
\vspace{-2.1cm}
\centering
\includegraphics[width=\textwidth]{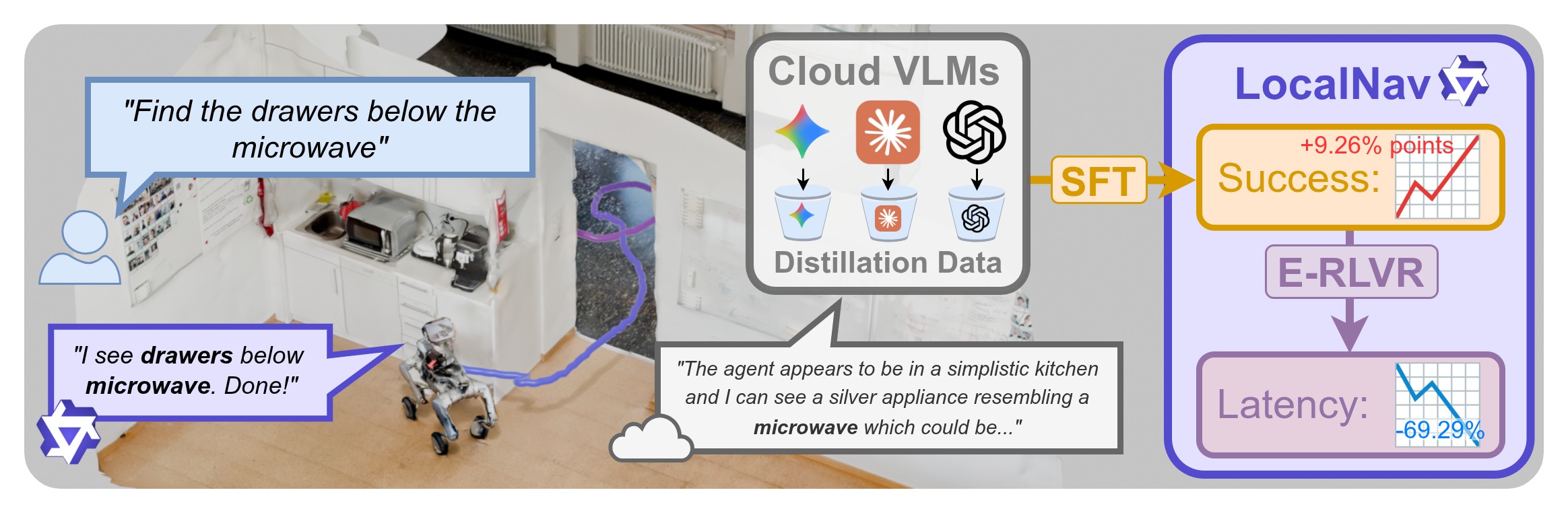}
\vspace{-0.75cm}
\captionof{figure}{Schematic overview of the proposed \textcolor{localnav}{LocalNav} system. \gls{sft} enables \gls{ogn} capabilities by distilling from large cloud frontier models, achieving a 9.26\%-point increase in \gls{sr} on a \emph{Qwen3.5-4B} model. \gls{erlvr} grounds the agent in its task and regularizes \gls{tg} length, decreasing generation latency by 69.29\%, and thus enabling on-board deployment on resource-constrained hardware.
}
\label{fig:title}
\vspace{-0.475cm}
\end{strip}
\glsresetall

\begin{abstract}
\glspl{vlm} have emerged in the robotic domain as a powerful tool that enables environmental perception with language context, serving as a catalyst for open-vocabulary tasks like \gls{ogn}. Yet, their computational footprint typically confines them to cloud execution, hindering low-latency inference with local deployment on resource-constrained robots. To address this challenge, we present a distillation strategy that aligns a lightweight, 4B-parameter local \gls{vlm} to our spatial-semantic navigation interface via frontier model traces for edge execution on embedded \acrshort{gpu} devices (e.g., \emph{Jetson Orin}). While cloud models establish a baseline achieving 41.61\% \gls{sr} on the \textit{HM3D OVON} benchmark, zero-shot deployment of our proposed architecture with \emph{Qwen3.5-4B} reaches only 25.97\% \gls{sr}.  We demonstrate that distilling frontier reasoning traces and subsequent finetuning with \gls{erlvr} obtains a 37.36\% \gls{sr}, substantially narrowing the gap to large cloud models. However, while deployable on memory-constrained devices, the model cannot compete with round-trip cloud inference time: we identify \acrfull{tg} as the main driver of latency and reduce it with \gls{tg}-regularized \gls{erlvr}. The combined method reduces latency by 69.29\% and retains a \gls{sr} of 35.53\%. Combined with quantization, this joint strategy yields a cumulative 80.21\% reduction in overall inference latency without significantly sacrificing performance, presenting a viable paradigm for local, low-latency \gls{vlm} execution on mobile robots.
\end{abstract}


\section{Introduction}
\IEEEPARstart{T}{he} emergence of \glspl{llm} and \glspl{vlm} with open-vocabulary capabilities \cite{achiam2023gpt, hurst2024gpt4o, gemini2.5, r1} is expanding embodied navigation to open-set instruction following \cite{uninavid, mtu3d, vlfm, openfrontier}. While closed-set agents have historically been limited to searching for generic categories defined during training (e.g., \textit{"cabinet"}), \glspl{vlm} provide the capacity for sophisticated scene understanding, which enables nuanced tasks such as locating \textit{"white cabinet beneath the microwave"}. Consequently, the agent reasons about semantically related objects or employs room-level navigation, utilizing its vision modality for visual verification. Formalized in open vocabulary benchmarks \cite{ovon, goatbench}, this open-set capability has spurred a paradigm shift in how foundation models are integrated into robotic systems. Following this transition, two distinct research trajectories have emerged:

\noindent \textbf{Low-Level Action Prediction:} Similar to recent advances in generalist manipulation policies \cite{openvla, pi0, pi0.5, smolvla, lee2025molmoact, dreamzero}, these methods build \gls{vla} models by conditioning pre-trained foundation models to predict low-level action tokens using \gls{bc}. While this grants the model direct control over the robot's low-level movement, it introduces several challenges. The discrete low-level action space $\mathcal{A}_\text{LL}$ used in \gls{ogn} is typically native to the \emph{Habitat} simulator, and, as such, can constrain the robot's agency. Secondly, the approach is heavily dependent on extensive \gls{sft} data for \gls{bc} and inherently constrains the system to the specific action spaces seen during training \cite{rt-2}. Crucially, as these models are trained end-to-end, fundamental robotic abilities such as obstacle avoidance and spatial memory must be implicitly learned \cite{uninavid}. Furthermore, because this architecture necessitates continuous inference to maintain control loops, typically requiring \SIrange{5}{10}{\hertz} for robot velocity control \cite{fan}, it imposes a significant computational burden, which presents a severe bottleneck for resource-constrained mobile robots \cite{vla_xpu}.\newline

\noindent \textbf{High-Level Decision-Making:} Rather than predicting low-level actions $\mathcal{A}_\text{LL}$, an emerging line of research utilizes \gls{vlm} foundation models strictly for abstract semantic reasoning over intermediate spatial representations, such as \glspl{sg} \cite{moma-llm, sg-nav}. By decoupling high-level reasoning from low-level control, the \gls{vlm} can focus on its core strengths: high-level scene comprehension and multimodal reasoning \cite{palm_saycan, l2r, rsspaper}. A dedicated low-level \emph{PointGoal} navigation policy can then pursue this goal \cite{dd-ppo}. This modularity enables the system to rely on dedicated navigation policies to safely traverse environments and avoid obstacles \cite{fan, dd-ppo, vlfm}. Furthermore, relying on structured representations, particularly \glspl{sg}, provides the agent with an explicit memory of the traversed environment. Because the \gls{vlm} is only invoked at sparse decision points, strict latency budgets are relaxed. However, \gls{sota} systems in this category typically rely on massive, cloud-bound \glspl{vlm} for abstract reasoning \cite{sg-nav, moma-llm, openfrontier}. Fully local deployment on mobile robot platforms remains a significant challenge due to the computational constraints of robotic hardware.

To summarize, low-level \gls{vla}-based systems necessitate continuous inference, while high-level decision-making approaches infer sparsely, but are cloud-bound to large frontier \glspl{vlm}. Hence, this work adopts a \textit{deployment-first} strategy, prioritizing the realization of \gls{ogn} on physical robotic hardware without relying on a cloud connection, as in \Cref{fig:title}. By aligning with a modular high-level decision-making paradigm, we enable competitive open-vocabulary \gls{ogn} while analyzing the efficacy of edge-deployable \glspl{vlm} and \gls{peft} strategies for embedded \gls{gpu}-based systems operating under strict latency, memory, and power budgets. We set the latency budget to match cloud round-trip times: a typical cloud query (\SI{\sim10}{\second}) corresponds to roughly 400 output tokens on a \textit{Jetson Orin AGX}, of which we target half, \num{\sim200} tokens, to leave headroom for concurrent on-board workloads, as in \Cref{sec:ablation}. Our core contributions are as follows:

\begin{enumerate}[label=\Roman*.]
    \item \textbf{Closing the Performance Gap to Cloud Models:} We distill \gls{ogn} capabilities from three frontier models into a locally deployable \textit{Qwen3.5-4B} \gls{vlm} and apply \gls{erlvr} on top of the best distilled student raising the \gls{sr} from 25.97\% to 37.36\%, and narrowing the gap to the best cloud model to 4.25 percentage points.
    \item \textbf{Token Generation Reduction Strategies:} Having identified autoregressive \gls{tg} as the dominant latency bottleneck, we investigate strategies to match cloud inference latency on a \textit{Jetson Orin AGX}. We compare zero-shot brevity prompting, \gls{sft} on shortened teacher traces, and a novel \gls{tg}-regularized \gls{erlvr}.
    \item \textbf{Cloud-Competitive On-Device Inference:} We find that \gls{erlvr} with \gls{tg} regularization yields the best trade-off between output length and navigation performance. Combined with quantization for faster \gls{tg}, our final model reduces total inference latency by 80.21\% relative to the \gls{sft} baseline while maintaining a \gls{sr} of 34.51\%.
\end{enumerate}

\section{Related Work}
\paragraph{Object Goal Navigation} Early approaches to \gls{ogn} generally fall into two categories: end-to-end learning or modular systems. End-to-end methods typically train policies using \gls{rl} in simulated environments to develop exploration behaviors \cite{dd-ppo, auxiliary_rl_ogn, scene_mem}. To then reduce dependency on training in simulation, modular pipelines construct intermediate representations, such as top-down semantic maps \cite{semexp, Naviformer, whatmattersinrl, t-diff}, explicit exploration frontiers \cite{ramakrishnan2022poni}, or \glspl{sg} \cite{HOZ_2021_ICCV, yang2019visual}, to guide the navigation policy. However, a fundamental limitation across these methods is their reliance on a closed set of object categories defined during training, preventing generalization to open-vocabulary goals or free-form instructions \cite{ovon, goatbench}.

\paragraph{Foundation Models for Navigation} To overcome these closed-set limitations, recent works leverage \glspl{vlm} either for direct low-level control or for high-level semantic reasoning. Methods focused on low-level control convert these models into \glspl{vla}, training them end-to-end to directly output discrete actions or short-horizon motion primitives \cite{navid, uninavid, navila, octonav, navFOM}. While compelling, \glspl{vla} require vast curated datasets \cite{navid, uninavid, navFOM}, lack interpretability \cite{ecot, firoozi2025foundation}, and remain too computationally heavy, even at 2-7B parameters, to run at the high control frequencies needed for closed-loop execution on embedded robot \glspl{gpu} \cite{vla_xpu}.
Rather than predicting low-level actions $\mathcal{A}_\text{LL}$, an alternative line of research leverages \glspl{vlm} strictly for high-level decision-making over intermediate spatial representations, such as value maps \cite{vlfm, intruct-nav}, \glspl{sg} \cite{sg-nav, moma-llm}, frontier-object maps \cite{fom-nav}, or set-of-marks prompting \cite{openfrontier}. Other works add structure via cognitive state machines \cite{cognav}, multi-sourced map fusion \cite{intruct-nav}, or query-based 3D grounding \cite{mtu3d}. By invoking the \gls{vlm} only at sparse decision points, these methods relax strict latency budgets, though \gls{sota} approaches \cite{vlfm, cognav, moma-llm} typically rely on cloud-hosted models (e.g., \textit{GPT-4o}), introducing network latency and cloud-connection dependencies.

Building on these high-level reasoning methods, our approach operates out-of-the-box with cloud models, and we show that the required reasoning capabilities can be distilled into a 4B-parameter model using a modest distillation dataset. As sparse high-level reasoning has been shown to tolerate cloud round-trip latency on physical robots~\cite{moma-llm, cognav}, we target matching this latency with fully on-board inference on a \textit{Jetson}-class device, eliminating the dependence on network connectivity.

\section{Methodology}\label{sec:method}

\subsection{System Architecture}\label{sec:architecture}
\begin{figure*}[!htb]
    \centering
    \vspace{-0.25cm}
    \includegraphics[trim={0 0.85cm 0 0.35cm},clip,width=\textwidth]{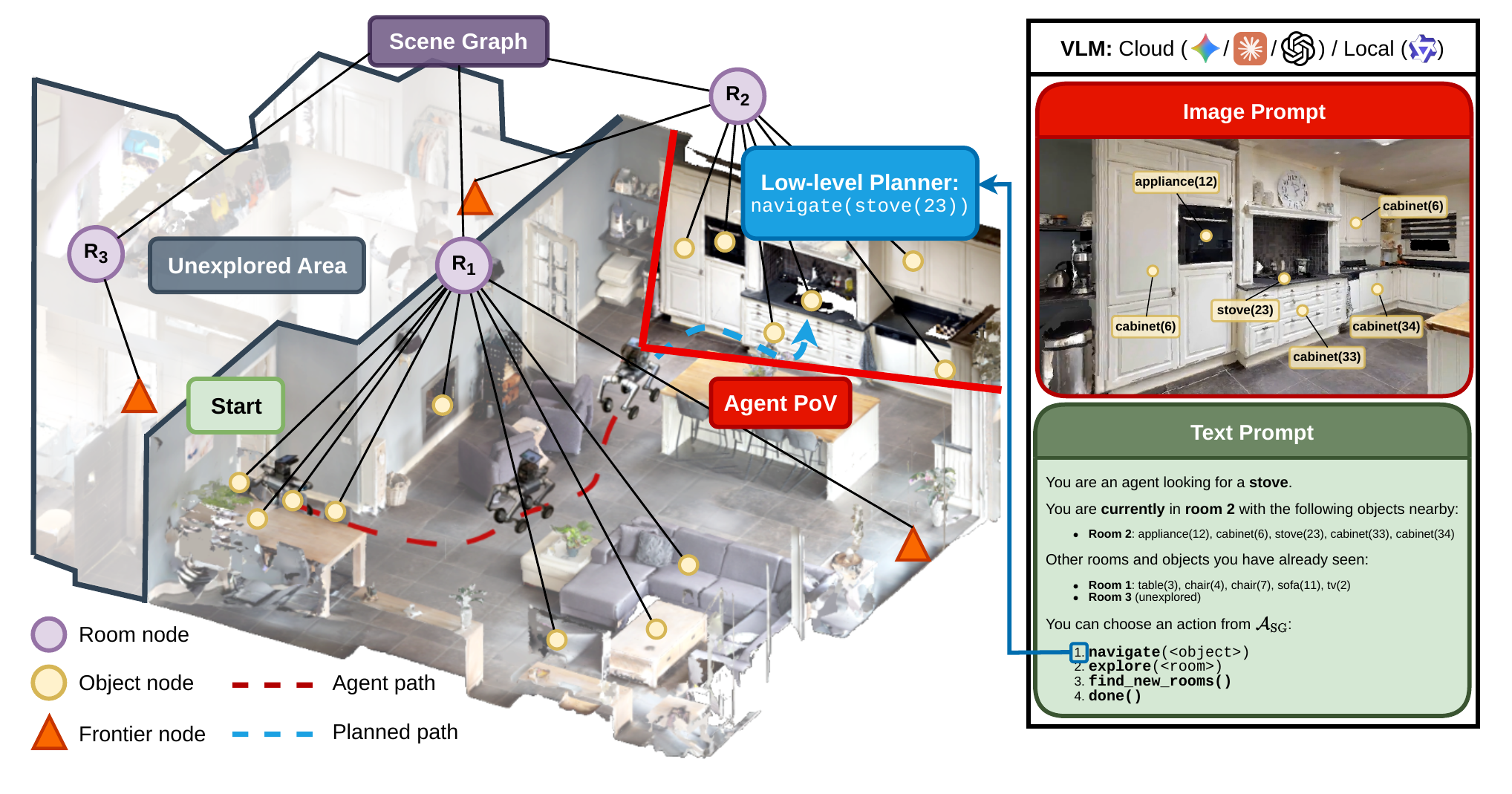}
    \caption{Overview of the system architecture. A 3D \gls{sg} is built in real time, providing information about \textcolor{purplei}{rooms}, \textcolor{yellowi}{objects}, and \textcolor{triangle}{unknown frontiers}. \gls{sg} \textcolor{yellowi}{objects} which are within the \textcolor{redi}{agent's PoV} are projected into the prompt image, while the overall state of the \gls{sg} is described in the simplified \textcolor{greeni}{text prompt} (full system prompt available in \Cref{fig:prompt_structure}). In this example, the \gls{vlm} chooses \texttt{navigate(stove(23))} as the next action. This invokes the low-level planner to navigate the agent to the 3D position of \texttt{stove(23)}.}
    \label{fig:system_arch}
    \vspace{-0.25cm}
\end{figure*}

\textcolor{localnav}{LocalNav} uses an explicit textual and visual representation of the scene and the agent's history. This choice has two benefits. First, framing the task as a text-and-image prompt aligns it with how instruct models are typically trained \cite{instructblip, gemma3,qwen3-VL}, which lets us obtain a competitive benchmark score of 41.6\% with cloud models such as \textit{Gemini, GPT} and \textit{Claude} \cite{gemini3, gpt5.4, claude4.6} out of the box, and allows us to finetune a small \gls{vlm} (\textit{Qwen3.5-4B}) with only modest distillation data $\sim$500 samples compared to multi-million \gls{sft} datasets \cite{uninavid, navFOM}. Furthermore, this modular abstraction provides a plug-and-play advantage, enabling drop-in replacements between different underlying \glspl{vlm}. Second, the explicit state representation makes the system more interpretable: failures can be attributed to specific components, such as \gls{sg} construction or \emph{PointGoal} navigation. \Cref{fig:system_arch} shows the modular design of our navigation agent: the \gls{vlm} makes high-level decisions from an explicit scene representation, while 
low-level navigation is delegated to a \emph{PointGoal} policy. The scene representation is an off-the-shelf \gls{sg} \cite{hughes2022hydra} constructed online from RGB-D and odometry, with semantic labels from ground-truth annotations in \textit{Habitat} and from \textit{Mask2Former} \cite{swin-mask2former} on the real robot. For \emph{PointGoal} navigation, we follow Yokoyama et al. \cite{vlfm} for simulation-, and Fan et al. \cite{fan} for robot deployment. Further details are specified in \Cref{subsec:local_planner}. At each decision step, \gls{sg} information is passed to the \gls{vlm} in two complementary ways. Firstly, the text prompt lists the agent's current location, explored and unexplored rooms, and observed objects with their IDs. Secondly, the image prompt is augmented by projecting the IDs of \gls{sg} objects that are visible in the frustrum of the agent, directly onto the RGB frame, in the spirit of set-of-mark prompting \cite{yang2023set}, grounding the symbolic IDs in concrete image regions. Given these inputs, the \gls{vlm} selects one of four actions $\mathcal{A}_\text{SG}$: \texttt{\{navigate(object\_id), explore(room\_id), find\_new\_rooms(), done()\}}. The first two invoke the local planner to navigate to the corresponding 3D location in the \gls{sg}; \texttt{find\_new\_rooms} drives the agent toward \gls{sg} frontiers, prioritizing those near doors and stairs, until a new room is added; \texttt{done} terminates the episode. Further information on the action space in Appendix~\ref{app:action_space}. 

\subsection{Local Planner}\label{subsec:local_planner}
During simulation in \emph{Habitat}, we deploy a standard, widely adopted \gls{ddppo} policy~\cite{dd-ppo, vlfm} to map the high-level pose commands generated by the agent into valid, discrete low-level actions ($\mathcal{A}_\text{LL}$), utilizing raw depth frames as its exteroceptive input. However, because this \gls{ddppo} baseline is optimized exclusively for simulation environments and limited to a discrete three-action space \texttt{\{forward, turn\_left, turn\_right\}}, it remains ill-suited for physical hardware deployment. 

To bridge this gap in real-world environments, we adopt the \gls{sru}-based point-navigation framework proposed by \cite{fan}. This framework takes the depth image from our onboard \emph{ZED2i} camera (see \Cref{subsec:handheld}) and computes continuous linear and angular velocity commands in $\{v_x, v_y, v_\omega\}$. The local planner utilizes a dedicated \gls{vae} to extract latent spatial representations that are inherently robust to real-world sensor noise, alongside an \gls{sru}-enhanced \gls{rnn} memory architecture for implicit spatial mapping. This enables real-world point navigation, with local obstacle avoidance and sub-goal pathfinding capabilities, allowing it to dynamically navigate toward target coordinates. For deployment, we modify and retrain the open-source implementation from \cite{fan} to match the robot embodiements utilised.

\subsection{SFT Data Generation} 
To transition to locally deployable models, we employ \emph{Qwen3.5-4B} as the backbone for our \gls{sg}-based navigation task. As shown in \Cref{tab:ablation}, the zero-shot performance of such a small model is insufficient, achieving only 25.97\%, compared to up to 41.6\% \gls{sr} of the cloud models. Consequently, the model must be trained specifically for this domain.
To achieve this, we record raw reasoning traces from \textit{Gemini 3.1 Pro, GPT-5.4,} and \textit{Claude Sonnet 4.6}, comprising image pairs and \gls{sg} states, generated directly by the aforementioned frontier models. A core advantage of this approach is its operational simplicity: it only requires logging the teacher model's outputs during task execution. Unlike alternative \gls{sft} strategies, our framework avoids complex feature-space alignment, the logging of intermediate embedding layers, or the meticulous tuning of multimodal pre-training data mixtures. Because this interaction occurs purely at the task level, our general system architecture remains entirely model-agnostic, allowing it to be seamlessly paired with any \gls{vlm}, whether a cloud-based frontier model or a locally deployed student.

To prevent data contamination, all training traces are gathered exclusively from the \textit{HM3D OVON} \texttt{train} set. Our data collection pipeline involves the cloud model performing the \gls{ogn} task in simulation using the privileged action space $\mathcal{A}_\text{priv}$ to separate \gls{vlm} capabilities from the performance of the low-level \emph{PointGoal} policy \cite{dd-ppo}, as elaborated in Appendix~\ref{app:action_space}; upon completion, only successful trajectories are selected for inclusion in the dataset. To investigate how the specific characteristics of different teacher models influence the student, we compile three distinct \gls{sft} datasets, one for each frontier model, and make them publicly available on \textit{HuggingFace}\footnote{Datasets: 
\href{https://huggingface.co/datasets/nibauman/objectnav-sft-gemini-3-1-pro-preview}{Gemini 3.1 Pro}, 
\href{https://huggingface.co/datasets/nibauman/objectnav-sft-gpt-5.4}{GPT-5.4}, 
\href{https://huggingface.co/datasets/nibauman/objectnav-sft-claude-sonnet-4.6}{Claude Sonnet 4.6}, 
\href{https://huggingface.co/datasets/nibauman/objectnav-sft-combined}{Combined}.}. Notably, each dataset contains only between 481 and 532 image-\gls{vlm}-output pairs. This dataset size is exceptionally small by contemporary finetuning standards, which can be in the order of millions of samples \cite{uninavid, navFOM}. An example dataset sample can be viewed in Appendix~\ref{app:prompt_and_response}. Because traces are collected exclusively from \texttt{train} while all reported results are evaluated on \texttt{val\_unseen}, and adaptation is performed via a rank-16 \gls{lora} with frozen vision layers (Appendix~\ref{app:definitions}), the student is prevented from overfitting to visual cues specific to the simulator.

\subsection{E-RLVR Closed-Loop Training Environment}\label{subsec:erlvr}
As analyzed in \Cref{subsec:compute}, autoregressive \gls{tg} is the primary bottleneck for execution latency. Therefore, minimizing the model's output length is critical for our \textit{deployment-first} principle. To compress these sequences while preserving the \gls{ogn} knowledge gained from \gls{sft}, we transition the model into an \gls{erlvr} closed-loop simulation environment. By allowing the \gls{vlm} to interact with the environment and updating its parameters via \gls{lora}-based \gls{grpo}, we shift from \textit{"textbook"} learning to \textit{"learning-by-doing"} \cite{boyle2025robotxr1}. By integrating an output length regularization reward, this active reinforcement incentivizes the model to internalize token-saving strategies without compromising navigation performance, a feat that cannot be achieved through zero-shot prompt engineering. As detailed in \Cref{sec:ablation}, zero-shot alternatives fail to balance these constraints: a concise \textit{caveman} style \cite{brussee2026caveman} yields negligible token savings, while suppressing \gls{cot} drastically degrades navigation performance. Static-length supervision, whether via truncated \gls{sft} traces or prompt-level brevity instructions, must fix a target length before the resulting action's consequence in the environment is known, decoupling brevity from task success. \gls{erlvr} instead rewards brevity only when the shortened response still advances the agent, since the rewards are computed from the same rolled-out trajectory and combined into a single advantage estimate, a coupling that is only possible because the verifier is the environment interaction itself, not a static label.

\begin{figure}[htb]
    \vspace{-0.0cm}
    \centering
    \includegraphics[trim={0 0 0 0.1cm},clip,width=1.0\linewidth]{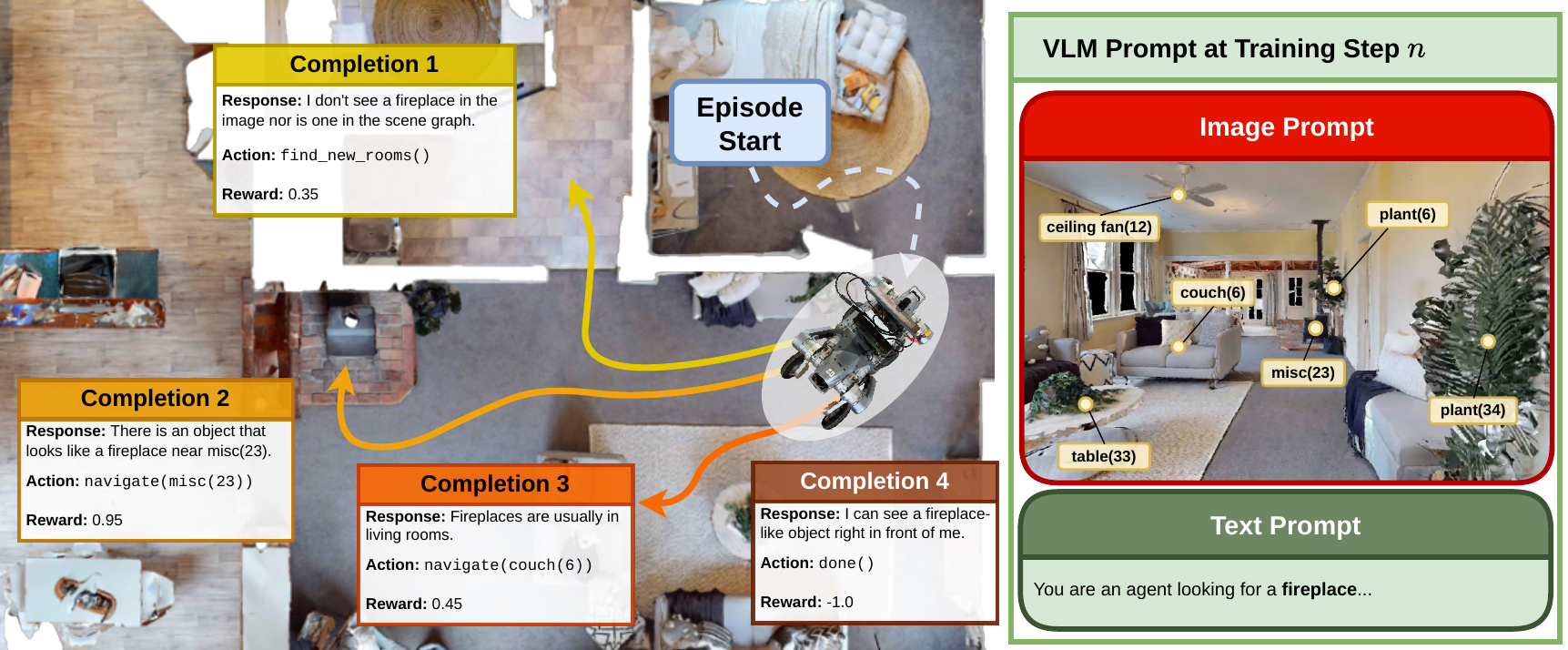}
    \caption{Schematic overview of the \gls{erlvr} training loop in \textit{Habitat}. The model samples four independent completions conditioned on the current camera observation and \gls{sg} state. Each action trajectory is rolled out individually in the simulator. Because the target (\emph{"fireplace"}) is already present within the \gls{sg} ($G \in SG$), the reward function penalizes distance-to-goal and verbosity ($R_{\text{brevity}}$), omitting exploration incentives.}
    \label{fig:erlvr}
    \vspace{-0.35cm}
\end{figure}

To execute \gls{erlvr} following \cite{boyle2025robotxr1, longnav-r1}, we integrate \gls{grpo} into our architecture of \Cref{sec:architecture}. As schematically illustrated in the training loop of \Cref{fig:erlvr}, the robot is initialized within a standard training episode. For each decision step, \gls{grpo} samples a group of $N$ independent \gls{vlm} completions (where $N=4$) from the same baseline prompt, conditioned on the active state of the \textit{Habitat} simulator and the current \gls{sg}. Each completion represents a distinct tool-call within the action space $\mathcal{A}_\text{SG}$. These $N$ streams fork isolated simulation branches that replicate the master environment and \gls{sg} states to execute the rollouts in parallel. Following execution, the resulting trajectories are evaluated and ranked via a multi-objective cumulative reward function defined as $R_{\text{tot}}=\sum_i R_i, \quad \forall i \in \{\text{done}, \text{nav}, \text{exp}, \text{brev}\}$. The single rewards are then defined as:
\vspace{-0.0cm}

{\small
\setlength{\arraycolsep}{2pt}
\begin{equation}\label{eq:reward_components}
\begin{aligned}
R_{\text{done}} &= \begin{cases} 1 & \text{success} \\ -1 & \text{fail} \\ 0 & \text{else} \end{cases}
& \;\;
R_{\text{nav}} &= \begin{cases} \overline{\Delta d} & G \in \text{SG} \\ 0 & \text{else} \end{cases} \\[4pt]
R_{\text{exp}} &= \begin{cases} 0 & G \in \text{SG} \\ \lambda_{\text{found}} & \text{found} \\ \overline{\Delta n} & \text{else} \end{cases}
& \;\;
R_{\text{brev}} &= \begin{cases} 1 & L \le L_{\text{t}} \\ f(L) & L_{\text{t}} < L < L_{\text{m}}. \\ -1 & L \ge L_{\text{m}} \end{cases}
\end{aligned}
\end{equation}
}
Each reward component is normalized, hence $R_i \in [-1, 1]$. The reward signal is composed of: $R_{\text{done}}$ penalizes or rewards termination depending on whether the agent stops within the success threshold of the goal, remaining neutral if an action other than $\mathcal{A}_{\text{done}}$ is selected; $R_{\text{nav}}$, rewarding progress ($\overline{\Delta d}$) towards the goal once localized; $R_{\text{exp}}$, incentivizing the discovery of new nodes ($\overline{\Delta n}$) or the goal itself entering the \gls{sg} ($G \in SG$);  and $R_{\text{brev}}$ enforces token efficiency by linearly rewarding the model from $-1$ to $1$ as the output length $L$ approaches the target length $L_{\text{t}}$, capping at $-1$ for any $L \geq L_{\text{max}}$. Here, $\overline{\Delta d} \in [-1, 1]$ and $\overline{\Delta n} \in [0, 1]$ represent the normalized progress toward the goal and the normalized count of newly discovered nodes, respectively. $C_\text{found}$ is a constant to reward the initial occurrence of $G \in SG$. Further terms are defined in Appendix~\ref{app:definitions}.

The cumulative rewards across all parallel trajectories are compared to determine local policy advantages. The highest-performing branch is selected to form the definitive starting state for the subsequent environment step. Then, \gls{grpo} leverages the performance variation across this local group to compute advantage-weighted gradients, updating the underlying \gls{lora} weights to optimize for navigation accuracy and token brevity simultaneously.

\subsection{Hardware Setup}\label{subsec:handheld}
\begin{figure}[htbp]
    \centering
    \begin{subfigure}[b]{0.4\columnwidth}
        \centering
        \includegraphics[width=\linewidth]{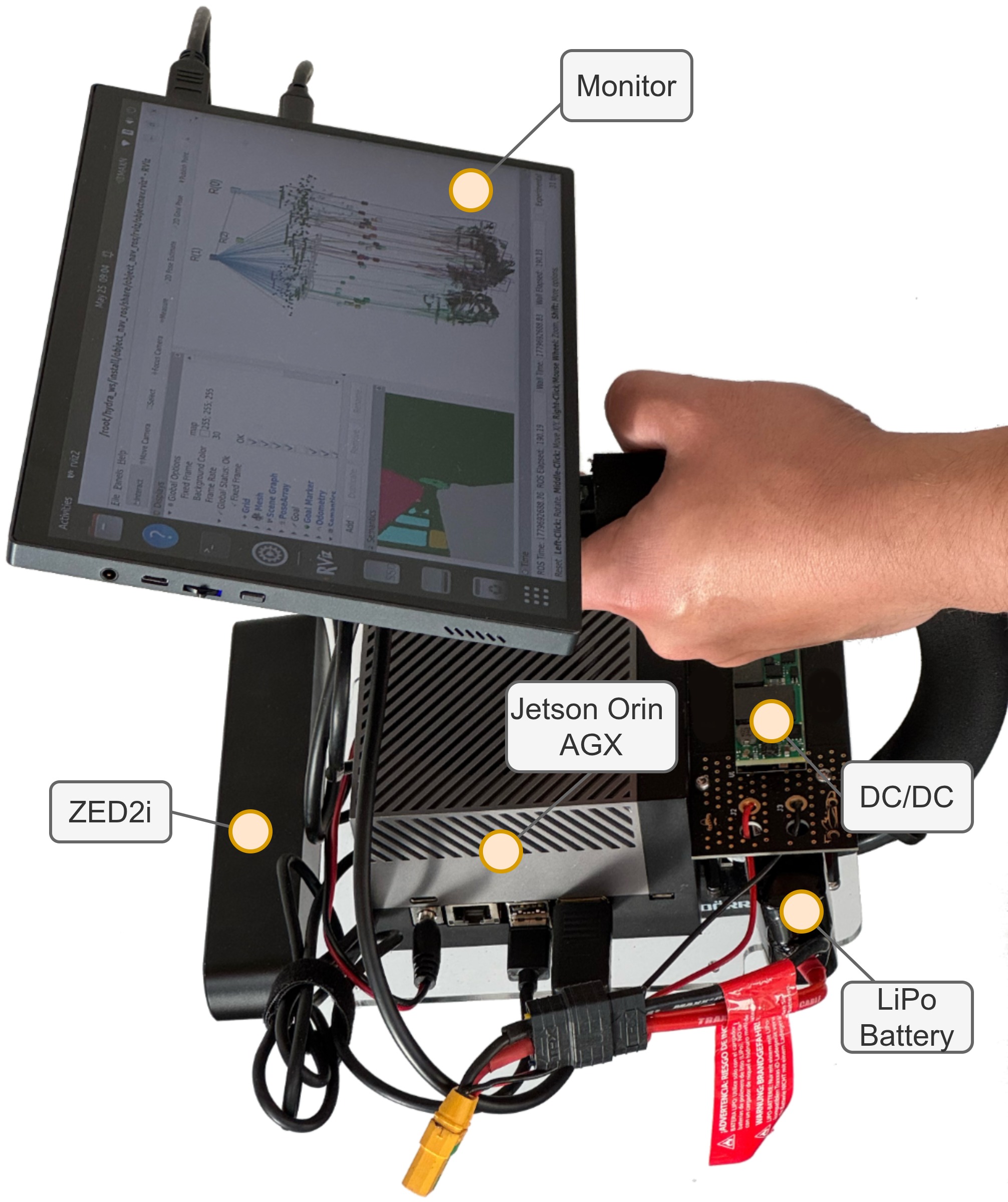}
        \caption{LocalNav Module}
        \label{fig:tartaruga}
    \end{subfigure}%
    \hfill
    \begin{subfigure}[b]{0.58\columnwidth}
        \centering
        \includegraphics[width=\linewidth]{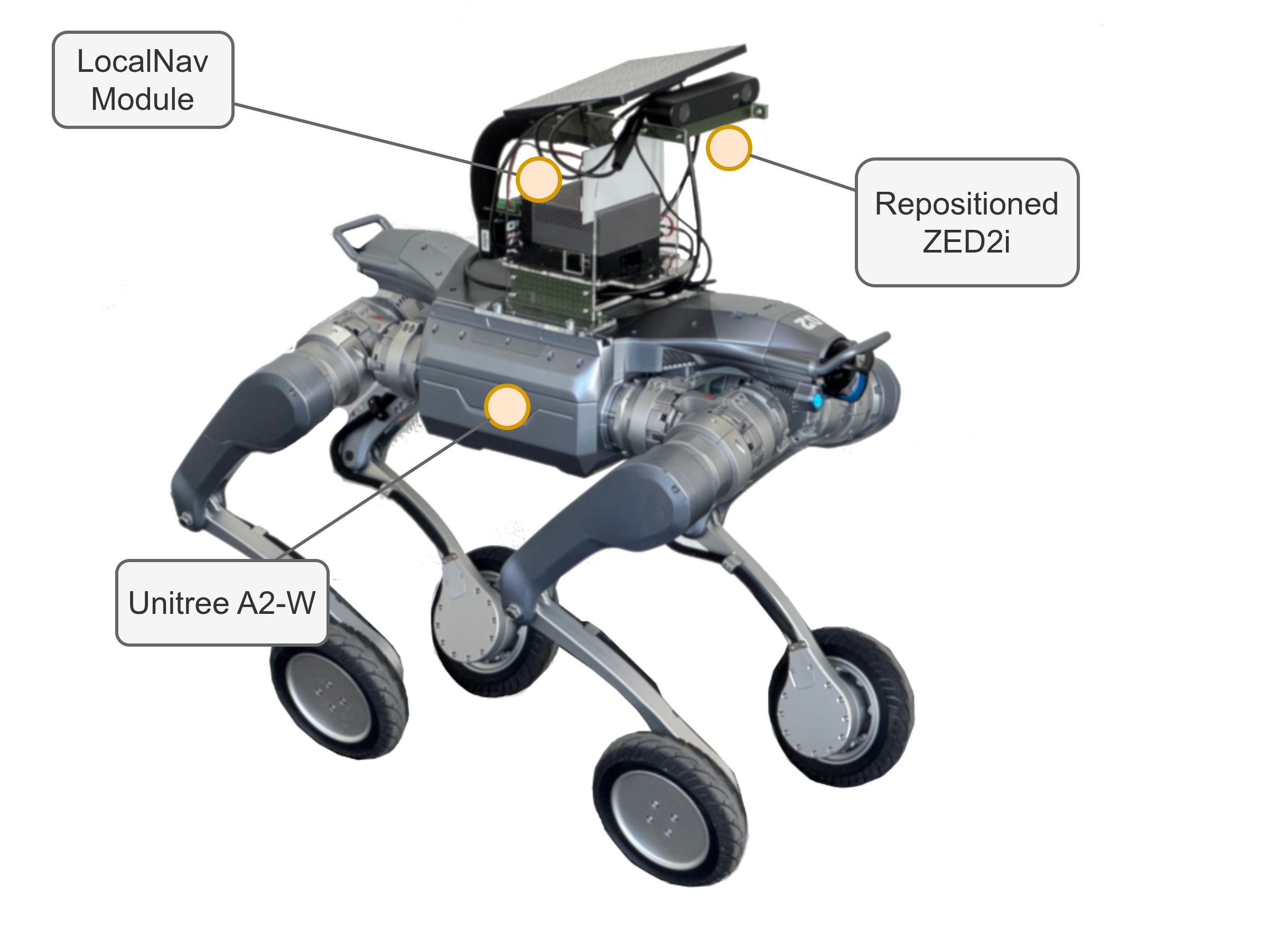}
        \caption{Unitree A2-W quadruped}
        \label{fig:quadruped}
    \end{subfigure}
    \caption{Experimental setup: (a) Depiction of the LocalNav Module with the \emph{Jetson Orin AGX}, \emph{ZED2i} stereo-depth camera, \emph{RViz} monitor, \gls{dcdc} converter, and 3S LiPo battery. (b) \textit{Unitree A2-W} robot with the LocalNav Module mounted and the \emph{ZED2i} repositioned for a clear \gls{fov}.}
    \label{fig:combined_setup}
\end{figure}

All real-world experiments use the \emph{LocalNav Module} (\Cref{fig:combined_setup}), a self-contained sensing and compute unit that runs the full navigation stack on board. It combines a \emph{Jetson Orin AGX}, a \emph{ZED2i} stereo camera, and a 3S LiPo battery with a \SI{120}{\watt}, \SI{12}{\volt} Murata\footnote{\href{https://pim.murata.com/asset/pim4/isolatedDCDCconverter/UWE-Q12_PDF_ISOLATEDDCDCCONVERTER?lastModifiedDatetime=20250707193436}{Murata UWE-Q12 Datasheet}} \gls{dcdc} converter. The \gls{sg} is built online by \emph{Hydra} \cite{hughes2022hydra} from the camera's RGB-D stream and \gls{vio} odometry. The module supports two configurations that differ only in how the $\mathcal{A}_\text{SG}$ pose goals are executed. In the handheld configuration (\Cref{fig:combined_setup}a), a human operator acts as the low-level planner and carries out the pose goals, allowing the perception and reasoning pipeline to be evaluated in arbitrary environments without a robot. In the robot-mounted configuration (\Cref{fig:combined_setup}b), the module is attached to a \emph{Unitree A2-W} quadruped, and the pose goals are passed to the PointGoal policy of \Cref{subsec:local_planner}, which converts depth images into velocity commands $\{v_x, v_y, v_\omega\}$ for fully autonomous navigation.

\section{Results}\label{sec:ablation}

Deploying a VLM-guided ObjectNav agent on-board hinges on two questions: can a locally deployable model approach the navigation performance of cloud frontier models, and can its inference latency match that of a cloud query? As a practical latency reference, we measure the round-trip time of 475 queries to \textit{Gemini~3.1~Pro}, the best-performing model in this work, over 100 episodes: the mean is 10.86\,s ($\sigma = 5.74$\,s, median 9.5\,s). We therefore take ${\sim}10$\,s per query as the bar at which on-board inference is comparable to the cloud. On an \textit{Nvidia Jetson Orin AGX} with \texttt{IQ4-XS} quantization, this corresponds to roughly 400 output tokens per query (\Cref{tab:compute}). Since the VLM shares the GPU with SG construction, segmentation, and the local planner during deployment, we target half this budget, ${\sim}200$ tokens, leaving headroom for concurrent on-board workloads.

\subsection{Navigation Performance Evaluation} All ablation experiments are evaluated on the full 3000-episode \textit{HM3D OVON} \texttt{val\_unseen} dataset using the low-level action space $\mathcal{A}_\text{LL}$, as in Appendix~\ref{app:action_space}. As detailed in \Cref{tab:ablation}, we distinguish between two primary types of metrics: \gls{ogn} metrics such as \gls{sr} and \gls{spl}, and deployment efficiency metrics: the average number of \gls{vlm} queries per episode, the frequency of redundant object re-visitations (\emph{Repetitions}), and the average output token length per query. The evaluation settings are defined in Appendix~\ref{app:eval_settings}.

\subsubsection{Comparison of Frontier Models} We evaluate the zero-shot performance of three different frontier models, \textit{GPT-5.4}, \textit{Claude Sonnet 4.6}, and \textit{Gemini 3.1 Pro}, seen on the first three rows of \Cref{tab:ablation}. \textit{Gemini 3.1 Pro} achieves the best navigation metrics (41.61\% \gls{sr}, 22.77\% \gls{spl}) and the lowest number of queries per episode (5.15), repetitions (0.0), and tokens per query (206.08).


\begin{table*}[!htb]
    \vspace{-0.0cm}
    \centering
    \resizebox{\textwidth}{!}{
    \begin{tabular}{l|c|c|c|c|c||c|c|c}
    \toprule
    \textbf{Model} & \textbf{SFT} & \textbf{RL} & \textbf{Prompt} & \textbf{SR [\%]} \bm{$\uparrow$} & \textbf{SPL [\%]} \bm{$\uparrow$} & \textbf{Queries [\#]} \bm{$\downarrow$} & \textbf{Repeat [\#]} \bm{$\downarrow$} & \textbf{Tokens [\#]} \bm{$\downarrow$} \\
    \toprule
    GPT-5.4 & \xmark & \xmark & normal & 38.03 & 19.12 & 7.61 & 0.02 & 285.62 \\
    Claude Sonnet 4.6 & \xmark & \xmark & normal & 39.7 & 19.72 & 6.464 & 0.008 & 606.40 \\
    Gemini 3.1 Pro & \xmark & \xmark & normal & \textbf{41.61} & \textbf{22.77} & \textbf{5.15} & \textbf{0.0} & \textbf{206.08}* \\
    \midrule \midrule
    Qwen3.5-4B & \xmark & \xmark & normal & 25.97 & 12.4 & 12.15 & 0.76 & 509.53 \\
    Qwen3.5-4B & Gemini & \xmark & normal & 30.63 & 15.06 & 9.06 & 0.41 & 245.24 \\
    Qwen3.5-4B & GPT & \xmark & normal & 28.63 & 14.25 & 7.26 & 0.17 & 275.56 \\
    Qwen3.5-4B & Claude & \xmark & normal & \underline{35.23} & \underline{17.77} & 7.41 & 0.12 & 486.71 \\
    Qwen3.5-4B & \{GPT/C/G\} & \xmark & normal & 30.37 & 15.16 & 7.36 & 0.15 & 443.13 \\
    \midrule
    Qwen3.5-4B & Claude & \xmark & caveman & 34.98 & 17.59 & 7.13 & 0.13 & 410.27 \\
    Qwen3.5-4B & Claude & \xmark & short-CoT & 34.2 & 16.75 & 8.41 & 0.3 & 261.38 \\
    \midrule
    Qwen3.5-4B & C-caveman & \xmark & caveman & 32.71 & 15.82 & \underline{6.65} & \underline{0.05} & 246.10 \\
    Qwen3.5-4B & C-short-CoT & \xmark & short-CoT & 31.01 & 15.89 & 7.29 & 0.23 & \underline{167.77} \\
    \midrule
    Qwen3.5-4B & Claude & \cmark \tiny{w/o $R_{brev}$}& normal & \textbf{37.36} & \textbf{18.01} & 8.76 & 0.31 & 424.20 \\
    Qwen3.5-4B & Claude & \cmark \tiny{w $R_{brev}$}& normal & 35.53 & 16.43 & \textbf{5.95} & \textbf{0.04} & \textbf{165.41} \\
    \bottomrule
    \end{tabular}
    }
    \caption{Ablation results for \gls{ogn} on the full \textit{HM3D OVON} \texttt{val\_unseen} split. Experiments use the $\mathcal{A}_\text{LL}$ action space, comparable to literature. \textbf{Queries} denotes average VLM calls per episode; \textbf{Repeat} tracks redundant object returns; \textbf{Tokens} represents average output length per VLM query. $^*$Reasoning tokens are not counted. \underline{Underlines} indicate the best metric in the class of distilled-only models.}
    \label{tab:ablation}
    \vspace{-0.25cm}
\end{table*}

\subsubsection{Improving Local Model Navigation Performance} The student baseline (\emph{Qwen3.5-4B}) performs poorly, achieving a \gls{sr} of only 25.97\% with the highest queries per episode of 12.15 and the a \gls{tg} of 509.53 on average. Applying targeted \gls{sft} using a modest corpus of teacher traces drastically improves \gls{ogn} performance across all metrics, though student capability remains heavily dependent on the distillation source. Among the individual single-teacher variants, the \emph{Claude}-distilled student achieves the highest performance with 35.23\% \gls{sr} and 17.77\% \gls{spl}, outperforming the \emph{Gemini}-distilled (30.63\% \gls{sr}, 15.06\% \gls{spl}) and \emph{GPT}-distilled (28.63\% \gls{sr}, 14.25\%) models.
Lastly, a mixed-teacher subset (\emph{\{GPT/C/G\}}, downsampled to $\sim$500 trajectories for data-size parity) yields lower performance (30.37\% \gls{sr}) than using the best teacher. Notably, distillation traces from the best-performing frontier model (\textit{Gemini 3.1 Pro}) do not lead to the best performing student (\textit{Qwen3.5-4B} distilled from \textit{Claude Sonnet 4.6}). Finally, we run \gls{erlvr} finetuning on the \textit{Claude}-distilled student without $R_{\text{brevity}}$ enforcement, resulting in the local model with the best navigation performance at 37.36\% \gls{sr} and 18.01\% \gls{spl}.

\subsection{Token Generation Runtime Analysis} 
\label{subsec:vlm_profiling}
\begin{table}
\centering
\begin{tabular}{lcc}
\toprule
\textbf{Metric} & \textbf{Prefill} & \textbf{Generation} \\
\midrule
Avg. tokens [tok]    & 2746.5 & 486.71 \\
Throughput [tok/s]   & 1271.4 & 18.04 \\
Runtime [s]          & 16.01 & 197.88 \\
\bottomrule
\end{tabular}
\caption{Prefill and \texorpdfstring{\gls{tg}}{TG} breakdown of average token counts, throughput, and latency of the \texorpdfstring{\gls{sft}}{SFT}-\emph{Claude} finetuned model on \emph{Jetson Orin AGX}.}
\label{tab:profile}
\end{table}

In order to gain a better understanding of how much the different model inference phases contribute to overall inference latency, \Cref{tab:profile} provides a quantitative profiling of the \gls{vlm}'s compute allocation where we evaluate the \gls{pp} (prefill) and \gls{tg} (decoding) execution phases for \textit{Qwen3.5-4B}. Performance metrics are compiled and averaged across a 100-episode evaluation run, exhibiting a mean of 7.41 \gls{vlm} invocations per episode. The profiling demonstrates that sequential \gls{tg} represents the primary execution bottleneck, demanding \SI{193.56}{\second} of the total \SI{213.89}{\second} average episode runtime ($\sim$92.5\% of end-to-end latency). These results indicate that edge deployment optimizations will be most effective when directly mitigating decoding latency either through reducing the number of tokens the model needs to decode or by speeding up the decoding itself.

\subsection{Inference Optimization via Prompting and Model Training}
Both the best-performing distilled model (486.71 tokens) as well as the overall best-performing model, finetuned with \gls{erlvr} (424.20 tokens), do not achieve the \gls{tg} target of ${\sim}200$ tokens per query. For this reason, and the findings of \Cref{subsec:vlm_profiling}, we investigate different strategies for reducing the number of generated tokens per query, while trying to maintain navigation performance.

\subsubsection{Prompting and SFT with Shortened Reasoning Traces}
Inference-only prompt modifications on the best-performing model (\emph{Claude}-SFT) show limited effectiveness, though task performance remains relatively stable, with \gls{sr} dropping only slightly (from 35.23\% to 34.98\% for \textit{caveman} and 34.20\% for \emph{short-CoT}). Yet, prompting with \textit{caveman} yields only a minor output token reduction (410.27 vs.\ 486.71). While \emph{short-CoT} prompting reduces generation length further (261.38 tokens), it degrades agent behavior: the average number of queries increases from 7.41 to 8.41, and redundant object re-visitations nearly triple (0.30 vs.\ 0.12).

Distilling new student models using shortened teacher traces is more successful. Training \textit{Qwen3.5-4B} with \textit{Claude} traces collected using \textit{caveman} and \textit{short-CoT} prompts reduces \gls{tg} to 246.10 and 167.77 tokens, respectively. While training on \textit{short-CoT} prompts is able to achieve sub-200 tokens per query, both of these approaches lead to a significant 2.52–4.22 \%-point drop in \gls{sr} compared to standard \emph{Claude}-\gls{sft} (32.71\% and 31.01\% vs. 35.23\%).

\subsubsection{Inference Optimization via \texorpdfstring{\gls{erlvr}}{ERLVR}} 
As an alternative strategy for \gls{tg} reduction, we add a brevity reward ($R_{\text{brevity}}$) to encourage shorter responses. Unlike the \gls{sft} approach, which strictly enforces a specific output length, finetuning with \gls{erlvr} allows the model to gradually adapt its response lengths to become shorter. Similar to \gls{sft} training with \textit{short-CoT} traces, this leads to sub-200 output tokens per query (424.20 vs. 165.41), but crucially, this token reduction comes at a smaller cost in navigation \gls{sr}. \gls{erlvr} with a brevity reward only drops 1.83 \%-points of \gls{sr} compared to \gls{erlvr} without \gls{tg} regularization, while also achieving the lowest number of queries (5.95) and repetitions (0.04) per episode. 

\begin{figure}[!htb]
    \centering
    \includegraphics[width=\columnwidth]{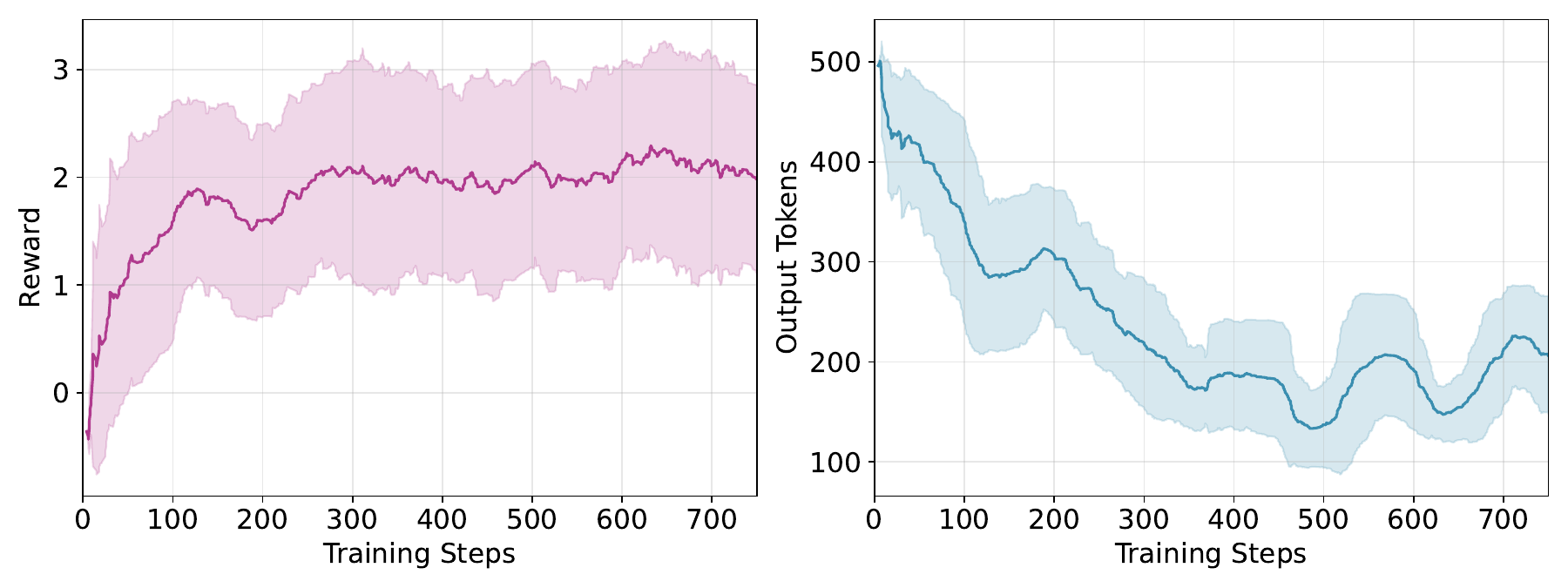}
    \caption{\glspl{erlvr} total cumulative reward on the left and mean token output length on the right during training. Both curves are smoothed with a rolling window of 100 samples. The shaded region represents one $\sigma$ of the rolling window.}
    \label{fig:rl_curves}
\end{figure}

\Cref{fig:rl_curves} depicts the cumulative reward and the mean token output length of a \gls{erlvr} training run using the \emph{Claude} \gls{sft} pretrained \emph{Qwen3.5-4B-SFT-Claude} model, aforementioned in \Cref{sec:ablation}. The run was performed with the hyperparameters as in Appendix~\ref{app:definitions}. As can be seen, the $R_\text{brevity}$ reward pushes the token output length from the initial $\sim$500 output tokens of the \gls{sft}-\emph{Claude} model, down to $\sim$200 tokens. Note, that during the \gls{erlvr} training, the \gls{vlm} is exposed to increased temperature for training diversity; hence the model behaviour during training does not perfectly represent the behaviour during evaluation, where the number of output tokens is even lower at $\sim$165 tokens.

\subsection{Inference Optimization via Quantization}\label{subsec:compute}
\begin{figure}[!htb]
    \centering
    \includegraphics[trim={0cm 0cm 0cm 0cm},clip,width=\columnwidth]{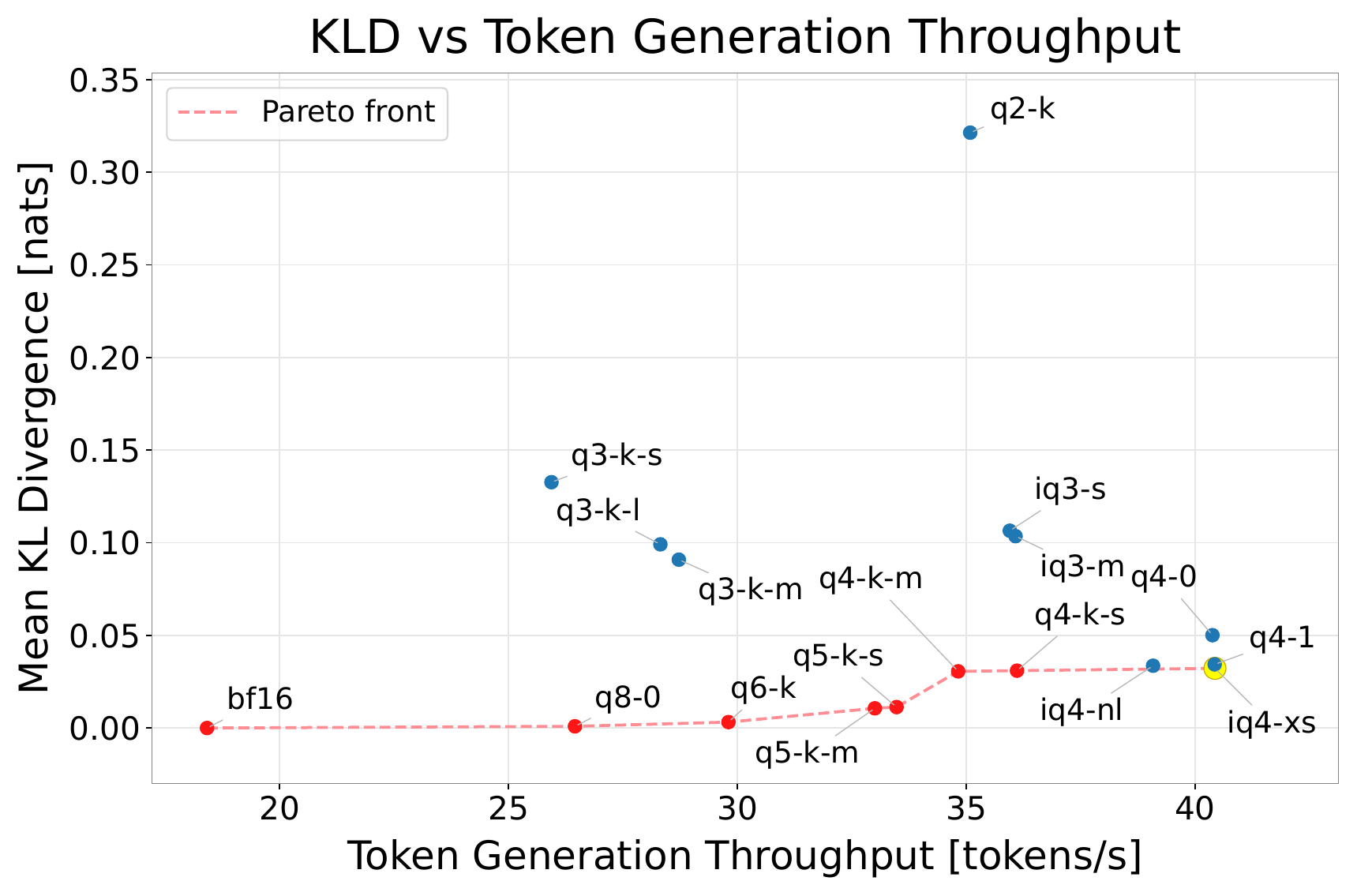}
    \caption{\gls{KLD} versus \gls{tg} throughput for all available quantization formats in \texttt{llama.cpp}. The quantization formats on the Pareto front are highlighted in red. The chosen \texttt{IQ4-XS} quantization is highlighted in yellow.}
    \label{fig:Throughputvskld}
\end{figure}

The findings in \Cref{subsec:vlm_profiling} showed that, besides reducing the number of tokens a model predicts, the most effective way to reduce inference latency is to improve the \gls{tg} speed of the model. To achieve a speedup in \gls{tg} we deploy all models using the highly optimized \texttt{llama.cpp} inference backend \cite{llamacpp} and evaluate all available quantization formats. To select the optimal quantization format for accelerating the overall runtime of the \gls{vlm} inference, we use \texttt{llama.cpp} to quantize the \gls{erlvr}-finetuned model into all supported data formats. As \Cref{subsec:vlm_profiling} showed that \gls{tg} dominates inference latency, we benchmark \gls{tg} throughput using real prompts from the navigation system for each quantization format to evaluate the runtime speedup each quantization scheme provides. To measure how well the language capability is preserved under each quantization format we evaluate the \gls{KLD} between the logit outputs of each quantized format and the native \texttt{BF16} format. \Cref{fig:Throughputvskld} shows the \gls{tg} throughput scales approximately proportionally with the reduction in data size along the Pareto frontier. This is because the \gls{tg} stage is memory bandwidth-bound. Reducing the model weight size through quantization improves memory bandwidth utilization and consequently increases overall compute utilization. As shown in \Cref{fig:Throughputvskld}, the \texttt{IQ4-XS} format provides the highest \gls{tg} throughput without significantly degrading the \gls{KLD}. Therefore, we select the \texttt{IQ4-XS} format for deployment.

\begin{table}
\centering
\scriptsize
\begin{tabular}{llccccc}
\toprule
\textbf{Model} & \textbf{Quant} & \makecell{\textbf{SR}$\uparrow$\\\textbf{[\%]}} & \makecell{\textbf{TG}$\uparrow$\\\textbf{[tok/s]}} & \makecell{\textbf{Avg. Out}$\downarrow$\\\textbf{[tok]}} & \makecell{\textbf{Queries}$\downarrow$\\\textbf{[\#]}} & \makecell{\textbf{Runtime}$\downarrow$\\\textbf{[s]}} \\
\midrule
SFT     & \texttt{BF16}   & 35.23  & 18.04 & 486.71 &  7.41 & 213.89 \\
SFT+RL  & \texttt{BF16}   & \textbf{35.53}  & 18.04 & \textbf{165.41} &  5.94 & 65.67 \\
SFT+RL  & \texttt{IQ4XS}  & 34.51  & \textbf{40.31} & 202.89 &  \textbf{5.91} & \textbf{42.32} \\
\bottomrule
\end{tabular}
\caption{Inference performance profile detailing the influence of quantization on \gls{sr}, \gls{tg} throughput, average output token count per query, average \gls{vlm} queries per episode, and average total latency per episode computed on a \textit{Jetson Orin AGX}.}
\label{tab:compute}
\end{table}

Finally, in \Cref{tab:compute} we can see, that at similar navigation \gls{sr} the model trained with \gls{tg}-regularized \gls{erlvr} produces 321.3 fewer output tokens and needs 1.47 fewer queries per episode than the best-performing SFT model. Running on an \textit{Nvidia Jetson Orin AGX} this result in a total inference runtime reduction from 213.89s to 65.67s per episode. Applying \texttt{IQ4XS} quantization on top of this model reduces navigation \gls{sr} by 1.02 percentage points but further reduces total runtime by 35.56\%.


\begin{figure*}[!htb]
    \vspace{-0.0cm}
    \centering
    \includegraphics[trim={0.0cm 0.6cm 14.5cm 2.45cm},clip,width=\textwidth]{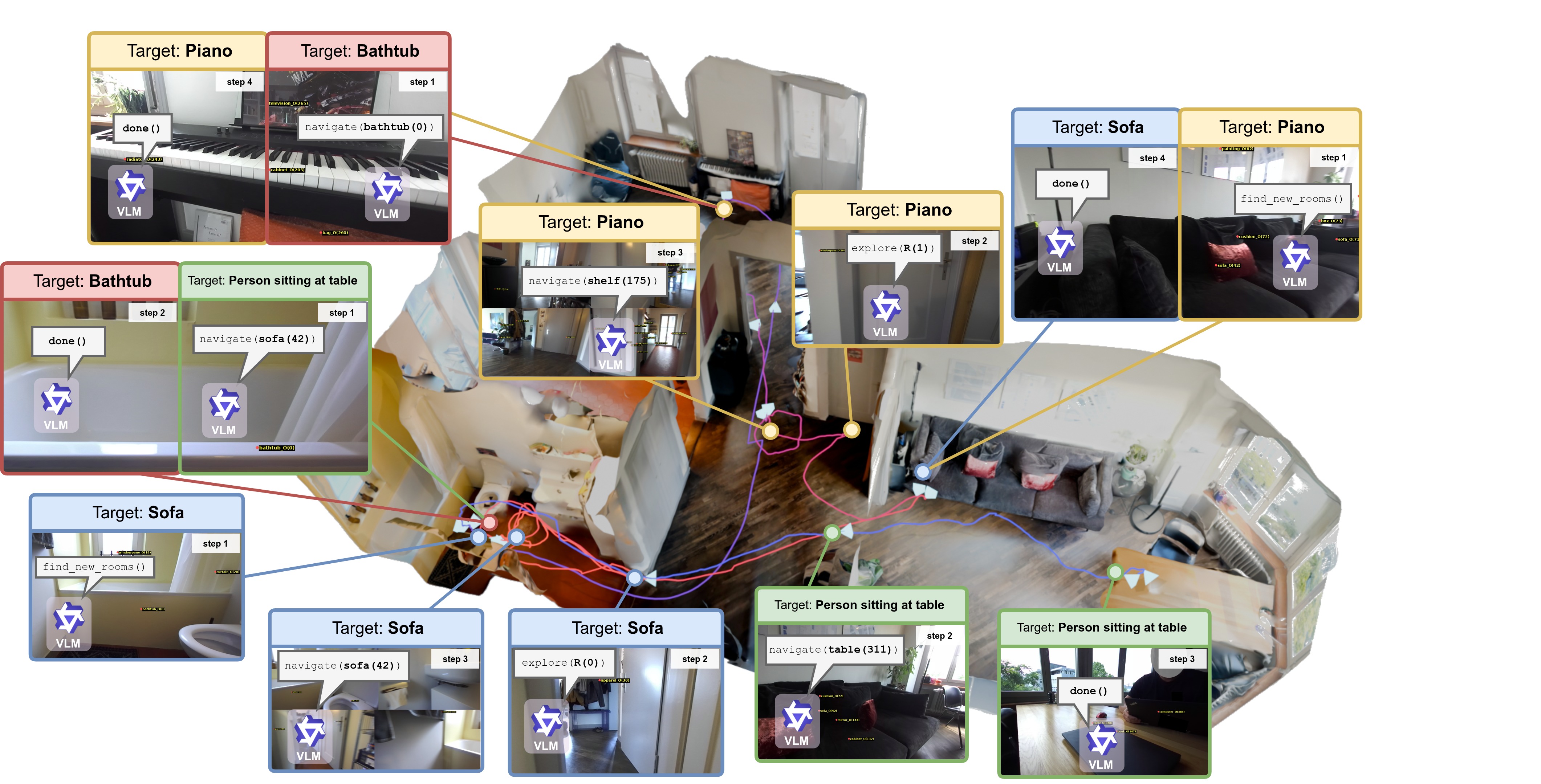}
    \caption{Real-world experiment conducted using the LocalNav Module in the handheld configuration (\Cref{subsec:handheld}). Starting in the bathroom, the agent is given four targets sequentially. Since this is a single deployment, the agent continuously builds up the \gls{sg}, reusing prior exploration for subsequent targets. We show the agent's PoV and response at each query, color-coded by target. The full trajectory is overlaid on the mesh, transitioning from \textcolor{redi}{red (start) } to \textcolor{localnav}{blue (end)}.}
    \label{fig:longstreet}
    \vspace{-0.25cm}
\end{figure*}

\subsection{Benchmark Evaluation}\label{subsec:comparison}
\begin{table}
\centering
\scriptsize
\begin{tabular}{l ccc cc}
\toprule
\textbf{Method} & \textbf{RGB} & \textbf{D} & \textbf{Sem} & \textbf{SR [\%]} & \textbf{SPL [\%]} \\
\midrule
BC \cite{ovon} & \cmark & \xmark & \xmark & 5.4 & 1.9 \\
DAgger \cite{ovon} & \cmark & \xmark & \xmark & 10.2 & 4.7 \\
RL \cite{ovon} & \cmark & \xmark & \xmark & 18.6 & 7.5 \\
DAgRL+OD \cite{ovon} & \cmark & \cmark & \xmark & 37.1 & 19.9 \\
VLFM \cite{vlfm} & \cmark & \cmark & \xmark & 35.2 & 19.6 \\
TANGO \cite{tango} & \cmark & \cmark & \xmark & 35.5 & 19.5 \\
MTU3D \cite{mtu3d} & \cmark & \cmark & \xmark & \textbf{40.8}$^*$ & 12.1 \\
Uni-NaVid \cite{uninavid} & \cmark & \xmark & \xmark & 39.5 & 19.8 \\
FOM-Nav \cite{fom-nav} & \cmark & \cmark & \xmark & 38.1 & \textbf{25.7} \\
OpenFrontier \cite{openfrontier} & \cmark & \cmark & \xmark & 39.0 & 20.1 \\
\midrule
Ours - Gemini 3.1 Pro & \cmark & \cmark & \cmark & \textbf{41.6} & \underline{22.8} \\
Ours - Qwen3.5-4B-Claude & \cmark & \cmark & \cmark & 37.3 & 18.0 \\
\bottomrule
\end{tabular}
\caption{\gls{ogn} performance benchmark on the \texttt{val\_unseen} split of \textit{HM3D OVON}. Reporting \gls{sr} and \gls{spl} of other works from their respective published results. \textbf{RGB}, \textbf{D} (Depth), and \textbf{Sem} (Semantics) indicate the sensor modalities used by each method. $^*$Best without semantic sensor.}
\label{tab:comparison}
\end{table}

Although the primary objective of this work is to establish a lightweight, hardware-deployable \gls{ogn} system for resource-constrained robotic platforms, we also compare our system to existing \gls{sota} methods on the \textit{HM3D OVON} benchmark. As shown in \Cref{tab:comparison}, our approach achieves performance competitive with existing \gls{ogn} work. Specifically, our pipeline utilizing the frontier cloud model \emph{Gemini 3.1 Pro} yields an \gls{sr} of 41.61\% and an \gls{spl} of 22.77\%. Our proposed training procedure allows the locally deployable 4B student model to remain competitive, achieving a \gls{sr} of 37.36\% and a \gls{spl} of 18.01\%. This performance gap, amounting to just 4.25 percentage points in \gls{sr} and 4.76 percentage points in \gls{spl}, demonstrates that local edge models can perform adequately against large frontier models for \gls{ogn}. As indicated in \Cref{tab:comparison}, our system relies on a semantic sensor to construct the \gls{sg}, consistent with common practice in the \gls{sg}-based navigation literature \cite{moma-llm, rana2023sayplan, ramakrishnan2022poni}. However, since these works \cite{moma-llm, rana2023sayplan, ramakrishnan2022poni} do not target the open-vocabulary \textit{HM3D OVON} benchmark, a direct comparison to them is not possible. Further details regarding metric definitions and exact simulation configurations are provided in Appendix~\ref{app:eval_settings}.

\subsection{Real World Experiments}\label{subsec:irl}
Finally, we evaluate our method in a real-world setting by deploying it on the LocalNav module as elaborated in \Cref{subsec:handheld} and completing 4 sequential \gls{ogn} tasks in a real apartment. \Cref{fig:longstreet} summarizes the run, showing the image prompts and \gls{vlm} responses at each step. Starting in the bathroom, the agent is asked to find, in order, a \textcolor{localnav}{"sofa"}, a \textcolor{yellowi}{"piano"}, a \textcolor{redi}{"bathtub"}, and a \textcolor{greeni}{"person sitting at a table"}. The first two queries require more steps, since the targets lie in unexplored regions and the agent must expand the \gls{sg} before it can navigate to them. The \textcolor{redi}{"bathtub"}, in contrast, is adjacent to the start location and already present in the \gls{sg} by the time it is queried, so the agent can leverage the explicit memory of the \gls{sg} without having to explore the space again. The final query illustrates open-vocabulary behavior: the agent navigates to a table and visually confirms that a person is seated at the table before terminating. 

\begin{figure}[!htb]
    \centering
    \includegraphics[width=\columnwidth]{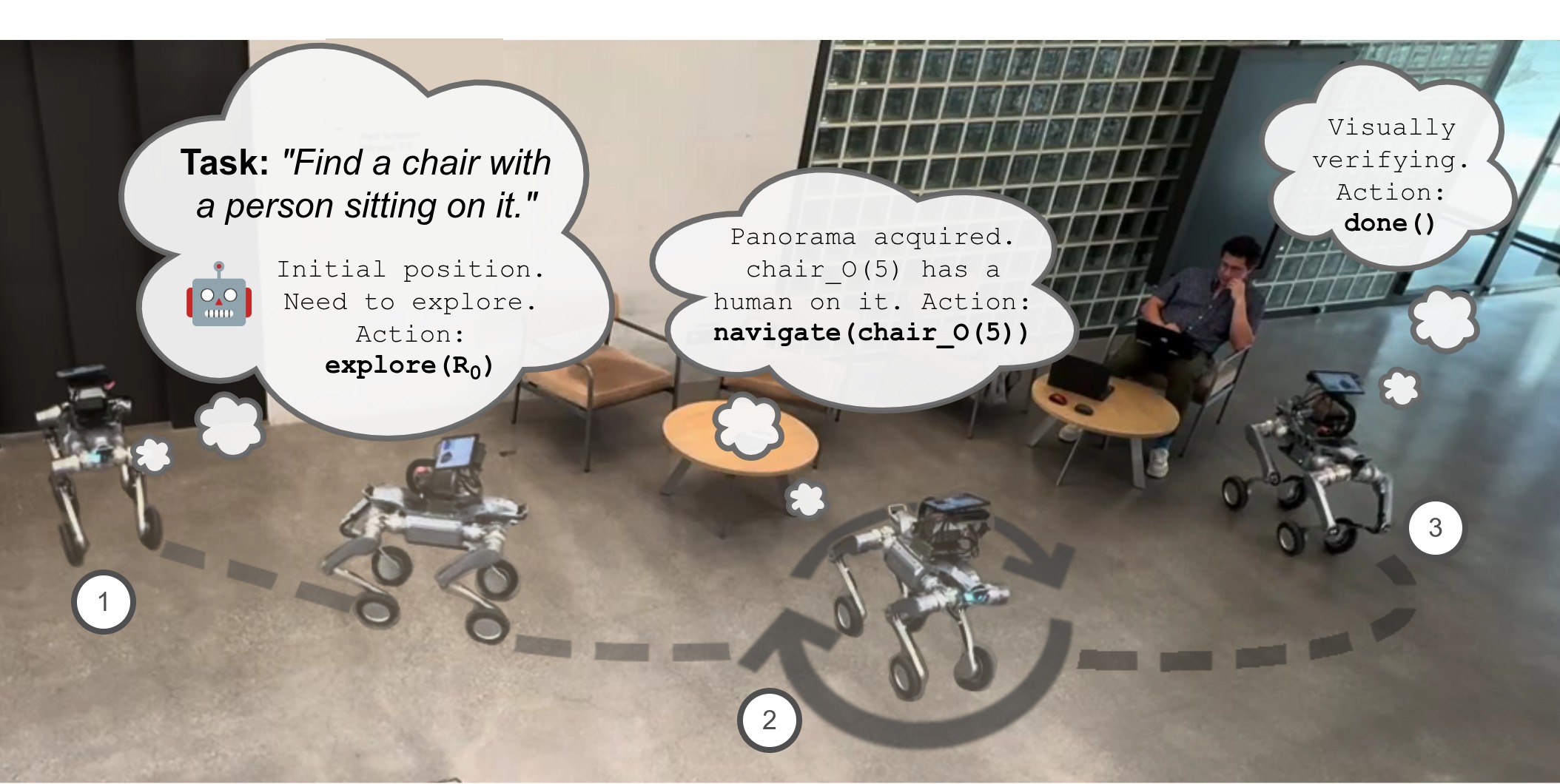}
    \caption{The proposed system mounted on a \emph{Unitree A2-W}. The open-vocabulary task is to navigate to a \emph{"chair with a person sitting on it."}. The robot first begins to explore its surroundings, as the goal is not in its immediate \gls{fov}. Once the goal has been spotted and anchored in the \gls{sg}, it can then navigate to it and visually verify it.}
    \label{fig:chair_ogn}
\end{figure}

Depicted in \Cref{fig:chair_ogn} is a further qualitative real-world demonstration of the proposed system performing an \gls{ogn} task on the \emph{Unitree A2-W} quadruped platform, to navigate to \emph{"a chair with a person sitting on it."} This represents an open-vocabulary task, as the \gls{vlm} must reason simultaneously over object categories and fine-grained attributes. 

In step 1, the robot is initialized without the goal object in its immediate \gls{fov}. It consequently decides to explore an available room in the \gls{sg} by executing the exploration action $\mathcal{A}^{\text{R}_0}_{\text{explore}}$. This action guides the robot toward the center of the room $R_0$. At step 2, the agent executes a full rotation to perform a lookaround behavior, acquiring a panoramic observation of the scene (a process that could alternatively be streamlined via a $360^{\circ}$ camera). This panoramic view enables the \gls{vlm} to detect multiple chairs and identify the specific instance matching the target attribute criteria. The agent then anchors this instance and executes a navigation action toward it via $\mathcal{A}^{\text{chair}_0}_{\text{navigate}}$. Upon arrival at step 3, the \gls{vlm} visually verifies the target configuration and terminates the episode.

\section{Conclusion}\label{sec:conclusion}
This work presents \textcolor{localnav}{LocalNav}, a framework enabling \gls{vlm} execution for \gls{ogn} on resource-constrained robotic platforms without dependence on a constant cloud connection. By developing a \gls{sg}-based architecture driven by frontier models, we establish a cloud-based baseline achieving a 41.6\% \gls{sr} on the \textit{HM3D OVON} benchmark. We demonstrate that with a modest training corpus, a lightweight locally deployable model can be aligned to perform \gls{ogn}, and that subsequent \gls{erlvr} raises its \gls{sr} to 37.4\%, narrowing the gap to cloud-based performance. Crucially, while \gls{sft} and \gls{erlvr} instill navigation capabilities, their verbose outputs prevent on-board inference from matching cloud latency. To address this, we augment \gls{erlvr} with \gls{tg} regularization, compressing output sequence lengths by 66.0\% while retaining a \gls{sr} of 35.5\%. Combined with quantization, this downstream optimization improves effective inference latency by 80.21\% without significantly deteriorating \gls{ogn} performance, presenting a viable paradigm for low-latency, embodied intelligence deployed natively on robotic hardware.

\section{Limitations}\label{sec:limitations}
\gls{sg}-based \gls{ogn} approaches are spatio-temporally limited. While the \gls{sg} updates between \gls{vlm} queries, the \gls{sg} cannot inherently bind transient, compositional semantics (e.g., distinguishing a generic \textit{"chair"} from a \textit{"chair with a person on it"}). Verifying these relations relies entirely on repetitive \gls{vlm} inference, which is only queried sparsely, and thus might miss the cues on the way. Hence, integrating a spatiotemporal scene graph framework such as \cite{daaaam} to track dynamic semantic attributes natively represents an interesting direction for future work.

\section{Acknowledgments}{We thank Dr. Franz Scherr for valuable technical discussions, as well as Maurice Brunner and Davide Plozza for their contributions to the robot hardware development and deployment. This work was supported by a grant from the Swiss National Supercomputing Centre (CSCS) under project ID lp160 on Alps and by the European Union's Horizon Europe research and innovation programme under Grant Agreement No. 101235536 (RADIANCE).}

\appendix
\subsection{Model Input and Output Example}\label{app:prompt_and_response}
\begin{figure*}[!htbp]
    \centering
    \includegraphics[width=\textwidth]{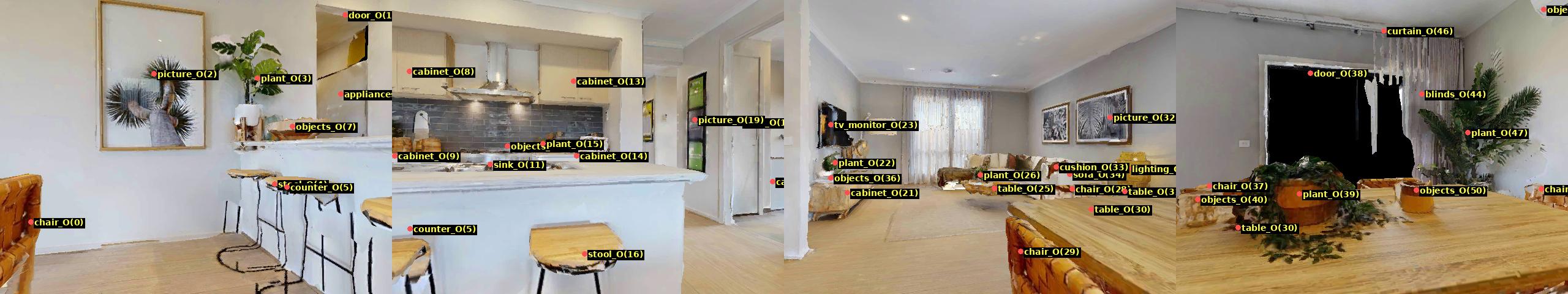}
    \caption{Panoramic image of an example \emph{Habitat} interaction, with the \gls{sg} objects projected into the image. This image corresponds to the \glspl{vlm} outputs in \Cref{tab:vlm_output_comparison}.}
    \label{fig:cloud_pano}

\vspace{1.5em} 

\begin{minipage}{\textwidth}
\centering
\scriptsize
\setlength{\tabcolsep}{4pt} 
\begin{tabularx}{\textwidth}{XXXX}
\toprule
\textbf{Claude-Sonnet-4.6} & \textbf{GPT-5.4} & \textbf{Gemini-3.1-Pro-Preview} & \textbf{ObjNav-Qwen3.5-4B-SFT-RL} \\ \toprule
575 tokens & 316 tokens & 1,497 tokens * & 104 tokens \\ \toprule
\textbf{Output Text:} & 
\textbf{Output Text:} & 
\textbf{Output Text:} & 
\textbf{Output Text:} \\
\texttt{<HISTORY>} \newline
No previous actions taken. This is the first step. \newline
\texttt{</HISTORY>} \newline
\texttt{<REASONING>} \newline
\textbf{Target Analysis:} \newline
- Air conditioners are commonly found in living rooms, bedrooms, or dining rooms - typically mounted on walls or ceilings. \newline
- The current room appears to be a combined kitchen/dining/living area based on the panorama... \newline
The \texttt{appliances\_O(6)} in R(0) is the most semantically plausible match for an air conditioner - "appliances" is a broad category that could include an AC unit. Let me navigate to it to investigate... \newline
\texttt{</REASONING>} & 

\texttt{<HISTORY>} \newline
No actions have been taken yet. I should start by using the initial panorama and scene graph to choose the most plausible room/object area for locating the air conditioner. \newline
\texttt{</HISTORY>} \newline
\texttt{<REASONING>} \newline
An air conditioner is most likely to be found in a living room, bedroom... The panorama shows an open-plan kitchen/living area... There is no explicit air conditioner label in the scene graph. The best next step is to explore the most plausible room in the scene graph that contains the living area, namely R(1)... \newline
\texttt{</REASONING>} & 

\texttt{<HISTORY>} \newline
This is the first step of the episode — no previous response. \newline
\texttt{</HISTORY>} \newline
\texttt{<REASONING>} \newline
The target is an air conditioner. Looking at the provided panorama, specifically the fourth view on the right, I can see a white, wall-mounted mini-split air conditioner unit situated high up on the wall above the window with blinds. The scene graph identifies the blinds as \texttt{blinds\_O(44)} and the room as R(1). To get within 1.0 meters of this air conditioner, I should navigate closer to the objects directly underneath it, such as \texttt{blinds\_O(44)}. \newline
\texttt{</REASONING>} & 

\texttt{<HISTORY>}\newline
None taken yet.\newline
\texttt{</HISTORY>}\newline
\texttt{<REASONING>}\newline
The scene graph shows R(0) has \texttt{appliances\_O(6)} which is a very strong match for an air conditioner. The room is R(0) with no assigned label, but it has appliances. Let me navigate to \texttt{appliances\_O(6)} to get visual confirmation.\newline
\texttt{</REASONING>} \\ \hline

\textbf{Action Output:} \newline
\texttt{navigate(6)} & 
\textbf{Action Output:} \newline
\texttt{explore(R(1))} & 
\textbf{Action Output:} \newline
\texttt{navigate(blinds\_O(44))} & 
\textbf{Action Output:} \newline
\texttt{navigate(appliances\_O(6))} \\ \hline
\end{tabularx}
\captionof{table}{Qualitative comparison of different \glspl{vlm} given the identical context, prompt and and input image \Cref{fig:cloud_pano}. While all model outputs yield valid and plausible actions, this comparison highlights the behavioral diversity across different models. Models are ordered from left to right: \emph{Claude Sonnet 4.6}, \emph{GPT-5.4}, \emph{Gemini-3.1 Pro Preview}, and \emph{Qwen3.5-4B} trained via \gls{sft} and \gls{erlvr}. Notably, \emph{Claude} generates the most verbose structural text, whereas \emph{Gemini}'s 1,497-token count *  is driven primarily by internal reasoning tokens omitted from the final visible block. All response blocks containing \emph{"..."} are shortened for viewing reasons.}
\label{tab:vlm_output_comparison}
\end{minipage}
\end{figure*}

This section aims to give an example of a \gls{vlm} input (system prompt and example \emph{Habitat} input image in \Cref{fig:prompt_structure} and \Cref{fig:cloud_pano}) and the diversity of model responses in \Cref{tab:vlm_output_comparison}. Furthermore, this prompt-image pair and \gls{vlm} response is an example from the open-sourced \gls{sft} combined dataset\footnote{Huggingface distillation combined dataset available at: \href{https://huggingface.co/datasets/nibauman/objectnav-sft-combined}{Combined Dataset}.}. 

\Cref{tab:vlm_output_comparison} shows a comparison between the different output styles of \glspl{vlm}. Given the context of \Cref{fig:cloud_pano} and the \gls{sg}-state in the prompt, all responses are equaly valid for this step. What can be seen, is the length of the output tokens, which is the largest for \emph{Claude-Sonnet-4.6} (\emph{Gemini-3.1-Pro-Preview} includes the reasoning tokens$^*$), as such, the proposed \emph{ObjNav-Qwen3.5-4B-SFT-RL} model, which was trained through \emph{Claude} generated \gls{sft} data, effectively reduces its token output length through the \gls{erlvr} stage.

\begin{figure}[!htbp]
\centering
\begin{minipage}{0.95\columnwidth}

\begin{tcolorbox}[
    colback=teal!5!white, colframe=teal!75!black, 
    title=\textbf{Task Definition \& Context}, 
    fonttitle=\bfseries, sharp corners, boxrule=1pt
]
\footnotesize
\textbf{\#\#\# ROLE} \\
You are an autonomous robot exploring an unknown indoor environment. You must integrate visual data from the camera with the provided Scene Graph to locate a target. \\[0.5em]
\textbf{\#\#\# OBJECTIVE} \\
Find and navigate to the closest: \textbf{air conditioner}
\end{tcolorbox}

\begin{tcolorbox}[
    colback=purple!5!white, colframe=purple!75!black, 
    title=\textbf{Spatial Context \& Multimodal Inputs (Scene Graph \& Camera)}, 
    fonttitle=\bfseries, sharp corners, boxrule=1pt
]
\footnotesize
\textbf{Current Location:} Unknown \\
\textbf{Objects within 1.0m:} \texttt{['chair\_O(45)', 'stool\_O(16)', ...]}\\
\textbf{\#\#\# ROOMS (Scene Graph)}
\begin{itemize}[leftmargin=1.5em, topsep=2pt, itemsep=1pt]
    \item \textbf{R(0):} \texttt{chair\_O(45), stool\_O(16), ...}
    \item \textbf{R(1):} \texttt{chair\_O(29), objects\_O(50), ...}
    \item \textbf{Unassigned:} \texttt{table\_O(30), plant\_O(39), ...}
\end{itemize}
\textbf{\#\#\# LOOKAROUND PANORAMA} \\
The image above is a stitched 360-degree panorama from your last lookaround scan... Cross-reference it with the Scene Graph...
\end{tcolorbox}

\begin{tcolorbox}[
    colback=orange!5!white, colframe=orange!75!black, 
    title=\textbf{System Constraints \& Executable Schema}, 
    fonttitle=\bfseries, sharp corners, boxrule=1pt
]
\footnotesize
\textbf{\#\#\# AVAILABLE ACTIONS}
\begin{itemize}[leftmargin=1.5em, topsep=2pt, itemsep=1pt]
    \item Action 0: \texttt{navigate}; Args: \texttt{[object\_id]}; Purpose: navigate to the specified object by its ID.
    \item Action 1: \texttt{explore}; Args: \texttt{[room\_id]}; Purpose: explores a room by navigating into the room and then giving a 360° scan of the specified room ID.
    \item Action 2: \texttt{done}; Args: \texttt{[]}; Purpose: call ONLY when you have found and navigated to the target. This terminates the task immediately.
    \item Action 3: \texttt{find\_new\_rooms}; Args: \texttt{[]}; Purpose: call this action to find new rooms that you have not seen before. This is used when you have explored your current surroundings and want to find new areas to explore.
\end{itemize}

\textbf{\#\#\# DECISION RULES}
\begin{itemize}[leftmargin=1.5em, topsep=2pt, itemsep=1pt]
    \item Use \texttt{find\_new\_rooms()} to find new rooms when all rooms in the current Scene Graph are exhausted.
    \item Use \texttt{explore(room\_id)} to get a panorama image of the room (Valid IDs: \texttt{\{'R(1)', 'R(0)'\}}).
    \item Call \texttt{done()} if you have visual confirmation that you are looking at an air conditioner within 1.0 meters.
\end{itemize}
\end{tcolorbox}

\begin{tcolorbox}[
    colback=red!5!white, colframe=red!75!black, 
    title=\textbf{Structured Response Format}, 
    fonttitle=\bfseries, sharp corners, boxrule=1pt
]
\footnotesize
\texttt{<HISTORY>} \\
Reflect on the history of your last actions, why you took them, and their outcomes. \\
\texttt{</HISTORY>} \\
\texttt{<REASONING>} \\
Answer target analysis questions. Justify your action based on the image and history. \\ \texttt{</REASONING>} \\
\texttt{<ACTION>\\
action\_name(args) \\
</ACTION>}
\end{tcolorbox}

\end{minipage}
\caption{Prompt of the proposed architecture, the content matches the input panorama image of \Cref{fig:cloud_pano} and the \gls{vlm} outputs of \Cref{tab:vlm_output_comparison}.}
\label{fig:prompt_structure}
\end{figure}

\subsection{Action Spaces}\label{app:action_space}
To isolate high-level visual-semantic reasoning from low-level execution errors, we define three distinct action spaces used across our architecture, planning frameworks, and evaluation configurations.

\paragraph{Low-Level Action Space ($\mathcal{A}_{\text{LL}}$)}
The standard discrete control action space provided by the \emph{Habitat} simulator for direct agent actuation:
\begin{align}
\mathcal{A}_{\text{LL}} = \{ & \texttt{forward}, \texttt{turn\_left}, \texttt{turn\_right}, \nonumber \\
& \texttt{look\_up}, \texttt{look\_down}\}
\end{align}

\paragraph{Semantic Scene Graph Action Space ($\mathcal{A}_{\text{SG}}$)}
The high-level action space output by the proposed \gls{vlm}, which operates on abstract semantic constructs and target coordinates:
\begin{align}
\mathcal{A}_{\text{SG}} = \{& \texttt{navigate(obj\_id)}, \texttt{explore(room\_id)}, \nonumber \\
& \texttt{find\_new\_room()}, \texttt{done()}\}
\end{align}
When deployed, actions sampled from $\mathcal{A}_{\text{SG}}$ produce pose goals. In simulation, these pose goals are translated into discrete $\mathcal{A}_{\text{LL}}$ commands via a \cite{dd-ppo} \emph{PointGoal} navigation policy. For physical deployments on real-world hardware, these coordinates are instead processed dynamically by the \cite{fan} planner for real-time obstacle avoidance and pathfinding.

\paragraph{Privileged Action Space ($\mathcal{A}_{\text{priv}}$)}
To completely decouple the high-level semantic reasoning capabilities of the \gls{vlm} from downstream tracking errors introduced by the low-level \emph{PointGoal} navigation policy, we define a privileged action space:
\begin{equation}
\mathcal{A}_{\text{priv}} = \mathcal{A}_{\text{SG}} \setminus \{\text{LL \emph{PointGoal} Execution}\}
\end{equation}
Under $\mathcal{A}_{\text{priv}}$, the underlying \gls{vlm} outputs the identical symbolic intent as defined in $\mathcal{A}_{\text{SG}}$. However, rather than executing iterative, low-level navigation steps via $\mathcal{A}_{\text{LL}}$, we use the built-in shortest path tool in \textit{Habitat} (relying on privileged simulator information) to compute the route from the agent's current location to the next $\mathcal{A}_{\text{SG}}$ target. We then subsample this path into \SI{0.5}{\metre} intervals and teleport the agent along these waypoints, allowing the \gls{sg} to be constructed incrementally. This control configuration provides an upper-bound performance baseline reflecting pure visual-semantic decision accuracy.

\subsection{Brief Prompt Styles}\label{app:prompt_styles}
To evaluate the limitations of zero-shot output-length reduction, we augment the standard prompt with two brevity-inducing instruction styles, applied at inference time on top of the \emph{Claude}-\gls{sft} model, and also used to collect the \emph{C-caveman} and \emph{C-short-CoT} \gls{sft} traces (\Cref{sec:ablation}).

\paragraph{Caveman} Following \cite{brussee2026caveman}, we append an instruction directing the \gls{vlm} to drop articles, auxiliary verbs, and other function words while retaining the required \texttt{<HISTORY>}, \texttt{<REASONING>}, and \texttt{<ACTION>} structure. This targets token savings at the linguistic surface level without altering the reasoning content itself.

\paragraph{short-CoT} This variant instructs the \gls{vlm} to omit the \texttt{<HISTORY>} and \texttt{<REASONING>} blocks entirely and respond with only the \texttt{<ACTION>} tag, suppressing the model's \gls{cot} before committing to an action. This targets token savings by removing the reasoning trace altogether, rather than compressing its language.

\subsection{Evaluation Settings}\label{app:eval_settings}
For comparisons against other methods on the full 3000 episodes of the \textit{HM3D OVON} \texttt{val unseen} benchmark (\Cref{tab:comparison}), the \gls{vlm} is tasked with predicting actions from $\mathcal{A}_{\text{SG}}$. When the \gls{vlm} issues a command such as \texttt{navigate(chair(14))}, the resulting navigation task is handled by a pre-trained \emph{PointGoal} navigator \cite{dd-ppo}. This navigator is provided with the 3D coordinates of the target (e.g., \texttt{chair(14)}), which are retrieved from the \gls{sg}. Because the \emph{PointGoal} navigator operates within the $\mathcal{A}_{\text{LL}}$ action space, our approach complies with standard \textit{Habitat}-based evaluation protocols, ensuring fair comparability between methods. The exact \textit{Habitat} parameters are detailed in Appendix~\ref{app:definitions}. We employ the standard \texttt{VIEW\_POINTS} success criterion, which considers an episode successful if the agent stops within a predefined distance of a viewpoint where an oracle can observe the target object. Following Chabal et al. \cite{fom-nav}, we set this distance threshold to \SI{0.25}{\metre}.

All results reported in this work, including \Cref{tab:compute} and \Cref{tab:ablation}, use the standard \SI{0.25}{\metre} \texttt{VIEW\_POINTS} threshold with the $\mathcal{A}_{\text{LL}}$ action space, consistent with the protocol used for \Cref{tab:comparison}. 

\subsection{Parameters and Additional Term Definitions} \label{app:definitions}
Sampling parameters used for \emph{Qwen3.5-4B} based models during evaluation:
\texttt{temperature}=0.7, \texttt{top\_p}=0.8, \texttt{top\_k}=20, \texttt{min\_p}=0.0, \texttt{repetition\_penalty}=1.0.
These are the suggested sampling parameters as in the \emph{HuggingFace} model page of \cite{qwen3.5}.

Referring to the linear scaling term of the brevity reward $R_\text{brevity}$ in \Cref{eq:reward_components}:
$$f(L) = 1 - 2 \frac{L - L_{\text{target}}}{L_{\text{max}} - L_{\text{target}}}$$

When $L = L_{\text{target}}$, reward is $1$. When $L \geq L_{\text{max}}$, reward is $-1$.

The cumulative reward of \Cref{subsec:erlvr}, is defined as $R = \sum\limits_{i \in \mathcal{I}} \lambda_i R_i$, with $\mathcal{I} = \{\text{done}, \text{nav}, \text{exp}, \text{brevity}\}$, where $\lambda_\text{done} = 2.0$, $\lambda_\text{nav} = 1.0$, $\lambda_\text{exp} = 1.0$, $\lambda_\text{brevity} = 0.25$.

\begin{table}[htbp]
\centering
\small
\setlength{\tabcolsep}{3pt}
\begin{tabular}{lc @{\hspace{1em}}c@{\hspace{1em}} lc}
\toprule
\textbf{Parameter} & \textbf{Value} & & \textbf{Parameter} & \textbf{Value} \\
\midrule
\midrule
\multicolumn{2}{l}{Hydra} & & \multicolumn{2}{l}{Habitat} \\
\midrule
Voxel size & \SI{0.05}{\meter} & & Resolution & $640{\times}480$ \\
Time step & \SI{0.5}{\second} & & Horizontal FOV & \ang{79} \\
Label space & MP-CAT40 & & Max depth & \SI{5.0}{\meter} \\
Frontier split scale & 0.5 & & Move forward & \SI{0.25}{\meter} \\
Min frontier voxels & 45 & & Turn angle & \ang{30} \\
Robot height & \SI{1.5}{\meter} & & Sliding & False \\
Modularity $\gamma$ & 2.0 & & & \\
\midrule
\multicolumn{2}{l}{\textit{\gls{sft}-\gls{lora}}} & & \multicolumn{2}{l}{\gls{erlvr}-\gls{lora} \gls{grpo}} \\
\midrule
\gls{lora} rank & 16 & & \gls{lora} rank & 16 \\
\gls{lora} $\alpha$ & 16 & & \gls{lora} $\alpha$ & 32 \\
Learning rate & $2{\times}10^{-5}$ & & Num. generations $N$ & 4 \\
Max steps & 500 & & KL penalty $\beta$ & 0.04 \\
Tune vision layers & False & & Learning rate & $2{\times}10^{-5}$ \\
& & & Max steps & 750 \\
& & & Rollout temp. & 1.0 \\
\bottomrule
\end{tabular}
\vspace{0.25cm}
\caption{Hyperparameter configurations for the \textit{Hydra} \gls{sg} module, \textit{Habitat} simulation environment, \gls{sft} distillation, and \gls{erlvr} training.}
\label{tab:hyperparameters}
\end{table}


\bibliographystyle{IEEEtran}
\bibliography{references}

\end{document}